\pdfoutput=1
\documentclass[10pt,journal,compsoc,onecolumn]{IEEEtran}
\ifCLASSOPTIONcompsoc
  \usepackage[nocompress]{cite}
\else
  \usepackage{cite}
\fi

\ifCLASSINFOpdf
\else
\fi

\usepackage[cmex10]{amsmath}
\usepackage{graphicx}

\usepackage{caption}
\usepackage{subcaption}
\usepackage{color}
\usepackage{transparent}

\DeclareMathOperator*{\divg}{div}

\DeclareMathOperator*{\argmin}{arg\,min}
\DeclareMathOperator*{\argmax}{arg\,max}

\begin{document}

% this should be in .bib file:
% @IEEEtranBSTCTL{IEEEexample:BSTcontrol,
%   CTLdash_repeated_names    = "no"
% }
\bstctlcite{IEEEexample:BSTcontrol}

\title{Multi-Atlas Segmentation using Partially Annotated Data: \\ Methods and Annotation Strategies}
% \title{Multi-Atlas Segmentation using Partially Annotated Data: a Unified Framework}


\author{Lisa~M.~Koch, 
  Martin Rajchl, 
  Wenjia Bai,
  Christian F. Baumgartner,
  Tong Tong,
  Jonathan Passerat-Palmbach,
  Paul Aljabar,
  and Daniel Rueckert}

% \institute{Biomedical Image Analysis Group, Imperial College London, UK \and Division of Imaging Sciences \& Biomedical Engineering, King's College London, UK}

% \markboth{IEEE Transactions on pattern analysis and machine intelligence}%
% {IEEE Transactions on pattern analysis and machine intelligence}

\IEEEtitleabstractindextext{%
\begin{abstract}

Multi-atlas segmentation is a widely used tool in medical image analysis, providing robust and accurate results by learning from annotated atlas datasets.  However, the availability of fully annotated atlas images for training is limited due to the time required for the labelling task.  Segmentation methods requiring only a proportion of each atlas image to be labelled could therefore reduce the workload on expert raters tasked with annotating atlas images. To address this issue, we first re-examine the labelling problem common in many existing approaches and formulate its solution in terms of  a Markov Random Field energy minimisation problem on a graph connecting atlases and the target image. This provides a unifying framework for multi-atlas segmentation. We then show how modifications in the graph configuration of the proposed framework enable the use of partially annotated atlas images and investigate different partial annotation strategies. The proposed method was evaluated on two Magnetic Resonance Imaging (MRI) datasets for hippocampal and cardiac segmentation. Experiments were performed aimed at (1) recreating existing segmentation techniques with the proposed framework and (2) demonstrating the potential of employing sparsely annotated atlas data for multi-atlas segmentation.

\end{abstract}

}

\maketitle

\IEEEdisplaynontitleabstractindextext

% \fi
\IEEEpeerreviewmaketitle

%%%%%%%%%%%%%%%%%%%%%%%%%%%%%%%%%%%%%%%%%%%%%%%%%%%%%%%%%%%%%%%%%%%%%%%%%%%%%%%%%%%%%%%%%%%%%%%%%%%
%%%%%%%%%%%%%%%%%%%%%%%%%%%%%%%%%%%%%%%%%%%%%%%%%%%%%%%%%%%%%%%%%%%%%%%%%%%%%%%%%%%%%%%%%%%%%%%%%%%
% *** INTRODUCTION ***
%%%%%%%%%%%%%%%%%%%%%%%%%%%%%%%%%%%%%%%%%%%%%%%%%%%%%%%%%%%%%%%%%%%%%%%%%%%%%%%%%%%%%%%%%%%%%%%%%%%
%%%%%%%%%%%%%%%%%%%%%%%%%%%%%%%%%%%%%%%%%%%%%%%%%%%%%%%%%%%%%%%%%%%%%%%%%%%%%%%%%%%%%%%%%%%%%%%%%%%

\IEEEraisesectionheading{\section{Introduction}\label{sec:introduction}}

\IEEEPARstart{I}{n} recent years, major efforts have been undertaken towards building large medical image databases such as ADNI \cite{Jack2008}. Segmenting anatomical structures in these images is often necessary to better understand physiological and pathological processes through quantitative analysis. As the wealth of data increases, manually annotating the images becomes prohibitive, especially for large 3D or 4D image datasets. Automated segmentation approaches may face challenges in large databases due to large variability in shape and appearance of the structures of interest, the presence of pathologies, or different imaging protocols used to acquire the images. 
In particular, it becomes increasingly desirable to develop robust and accurate segmentation techniques that rely on minimal manual input or weak supervision.

Multi-atlas segmentation \cite{Heckemann2006multiatlas,Rohlfing2004,Klein2005multiatlas} has proven to be a successful and robust tool and is widely used in the medical imaging community \cite{Iglesias2015a}. The approach generally relies on label propagation from multiple atlases (i.e. fully annotated training images) to a target image. 
Using multiple atlases offers the important advantage of capturing anatomical variability.
Ideally, the atlases should match the population to be segmented \cite{Wolz2010}. However, suitable atlases are not always available for large image databases, especially if the images in the database exhibit large variabilities, e.g. due to the presence of disease or aging processes. This motivates the use of training data obtained with different annotation strategies, where atlas images are only \emph{partially} annotated, drastically reducing the labelling effort per image and therefore allowing expert raters to (partially) annotate more training images in the same time. To employ partially annotated atlas data while building on the success of multi-atlas segmentation, we propose a generalisation of the labelling problem in existing multi-atlas segmentation methods. In the following paragraphs, we review relevant work in the field before identifying the main contributions of this paper.

Many multi-atlas segmentation techniques use non-linear registration to warp segmentations from multiple suitable atlases to a target image~\cite{Heckemann2006multiatlas,Artaechevarria2009,Aljabar2009atlasselection,Rohlfing2004,Klein2005multiatlas,Sabuncu2010}. The target segmentation can be formed by fusion of the propagated labels, for example by applying a majority vote rule \cite{Heckemann2006multiatlas,Aljabar2009atlasselection} or another combination strategy such as a weighted average based on global or local similarity measures between the target and atlas images \cite{Artaechevarria2009,Bai2013}. In \cite{Sabuncu2010}, a probabilistic framework was presented where the above-mentioned vote rules are expressed with a generative label fusion model.
This was extended in \cite{Bai2013} to incorporate non-local label fusion and registration uncertainty, and in \cite{Iglesias2015} to allow the use of atlases annotated with different labelling protocols.
Other combination strategies include STAPLE \cite{Warfield2004}, where label fusion weights are estimated with an expectation-maximisation algorithm, or Joint Label Fusion \cite{Wang2012}, where correlations among atlases are taken into account. To account for high local anatomical variability between images, and to relax the requirement for accurate registration, patch-based segmentation\,\cite{Coupe2011PatchBased,Rousseau2011} has been introduced. Using this approach, the label fusion step employs a non-local weighted average of voxel labels in a small neighbourhood of the atlas images, with weights based on the similarities of patches centred on the compared voxels.

Considerable improvements in segmentation accuracy can be achieved by using the label propagation results as prior probabilities in subsequent refinement steps, combining them with regularisation terms and an intensity model in a Markov Random Field (MRF) formulation~\cite{VanderLijn2008,Lotjonen2010,Makropoulos2014,Ledig2015,Rajchl}. This was first suggested by \cite{VanderLijn2008} in combination with graph-cuts~\cite{Boykov2001}, whereas~\cite{Lotjonen2010} proposed an expectation-maximisation approach, which was also adopted in \cite{Makropoulos2014} and \cite{Ledig2015}.

All of the above methods rely on the availability of a fully annotated atlas dataset with the aim to segment an individual target image. It has been shown that, in general, segmentation accuracy decreases when fewer \cite{Heckemann2006multiatlas} or less similar \cite{Aljabar2009atlasselection} atlases are used. However, segmentation methods requiring fewer atlases (i.e. training data) while preserving accuracy are highly desirable, as they could reduce the workload of raters who manually annotate these atlases.
Recently, a number of methods have been proposed for iterative label propagation, which allow labels from a small set of annotated atlas images to be propagated to similar images or image regions in the test population~\cite{Koch2014,Wolz2010,kuettel2012imagnet,Rubinstein2012,Cardoso2015}. These methods avoid error-prone registration between dissimilar images by only propagating information between similar images which are easy to register. They therefore exploit the unlabelled test population in a semi-supervised learning setup and thus reduce the amount of labelled atlas data necessary to achieve accurate segmentation results.

Other strategies to reduce the manual workload that have been proposed in the computer vision and medical imaging community employ weak supervision. This includes annotations in the form of bounding boxes around an object instead of pixel-wise labelling, such as proposed in GrabCut \cite{Rother2004} and recently extended to 3D bounding boxes in \cite{Chen2014}, scribbles that only annotate part of an image (e.g.~\cite{Boykov2000}), or image tags which only describe which class is present in an image (e.g.~\cite{Xu2014}). \cite{Xu2015} give a good summary of the various forms of weak supervision and propose a unified framework for segmentation in computer vision datasets. In the context of multi-atlas segmentation, \cite{Landman2012} proposed a modification of the STAPLE algorithm~\cite{Warfield2004} that can deal with missing annotations in the atlas images. 

A frequently used method to efficiently solve the labelling problem is to express it as a Markov Random Field (MRF) energy function~\cite{Li1994} and minimise it using min-cut/max-flow techniques~\cite{Boykov2000,Boykov2001,Yuan2010,Yuan2010a}. The MRF is normally defined by a graph constructed on a regular grid that represents the target image. However, some applications formulate an MRF energy function on graphs connecting \textit{multiple} images.  Recently,~\cite{Han2011} applied graph-cuts for co-segmentation of pairs of PET and CT images by minimising an MRF energy function which penalises tumour segmentation differences between a PET and CT image of the same subject. \cite{Trus2014} used an extension of continuous max-flow \cite{Yuan2010} for simultaneous prostate segmentation in multiple 2D slices while penalising segmentation differences between slices. Continuous max-flow (CMF) solves the continuous counterpart to the discrete min-cut/max-flow problem~\cite{Yuan2010} and it can be computed using a reliable, inherently parallelisable multiplier-based algorithm with guaranteed convergence. This makes it suitable for the optimisation of large labelling problems.

%%%%%%%%%%%%%%%%%%%%%%%%%%%%%%%%%%%%%%%%%%%%%%%%%%%%%%%%%%%%%%%%%%%%%%%%%%%%%%%%%%%%%%%%%%%%%%%%%%%
% *** Our contribution ***
%%%%%%%%%%%%%%%%%%%%%%%%%%%%%%%%%%%%%%%%%%%%%%%%%%%%%%%%%%%%%%%%%%%%%%%%%%%%%%%%%%%%%%%%%%%%%%%%%%%
\subsection*{Our contribution}

In this paper, we propose methods and annotation strategies which enable the use of partially annotated data for multi-atlas segmentation, with the main goal of reducing the required manual labelling effort. As a first contribution, we propose a unifying framework for multi-atlas segmentation using a novel graphical representation of the labelling problem. In Sec.~\ref{sec:multiatlas} we demonstrate how label fusion, spatial regularisation, and data models can be expressed simultaneously using this representation. To optimise the arising MRF energy function, we provide an efficient optimisation scheme based on continuous max-flow~\cite{Yuan2010,Yuan2010a}.

We then show in Sec.~\ref{sec:segmentation_using_partially_annotated_atlas_data} how the proposed framework can be used to go beyond the abilities of existing multi-atlas segmentation techniques:
The proposed flexible graph structure allows a relaxation of the annotation requirements in atlas images. This means that our framework naturally allows the use of atlases that were only partially annotated, resulting in a reduced manual labelling effort for expert raters.

In Sec.~\ref{sec:segmentation_using_partially_annotated_atlas_data} we examine different partial
annotation strategies and investigate modifications in the graph configuration to optimally exploit
partially annotated atlas data in the segmentation process.
Experiments on hippocampal (Sec.~\ref{sec:exp-unified-framework} and \ref{sec:exp-pa-slices}) and
cardiac segmentation (Sec.~\ref{sec:exp-pa-scribbles}) highlight the performance of the proposed
framework and shed light on some of the possibilities it offers for employing partial annotations such as missing slices or scribbles.
A preliminary version of this work was presented in~\cite{Koch2015}.

%%%%%%%%%%%%%%%%%%%%%%%%%%%%%%%%%%%%%%%%%%%%%%%%%%%%%%%%%%%%%%%%%%%%%%%%%%%%%%%%%%%%%%%%%%%%%%%%%%%
%%%%%%%%%%%%%%%%%%%%%%%%%%%%%%%%%%%%%%%%%%%%%%%%%%%%%%%%%%%%%%%%%%%%%%%%%%%%%%%%%%%%%%%%%%%%%%%%%%%
% *** UNIFIED MULTI-ATLAS SEGMENTATION ***
%%%%%%%%%%%%%%%%%%%%%%%%%%%%%%%%%%%%%%%%%%%%%%%%%%%%%%%%%%%%%%%%%%%%%%%%%%%%%%%%%%%%%%%%%%%%%%%%%%%
%%%%%%%%%%%%%%%%%%%%%%%%%%%%%%%%%%%%%%%%%%%%%%%%%%%%%%%%%%%%%%%%%%%%%%%%%%%%%%%%%%%%%%%%%%%%%%%%%%%

\section{Unified Framework for Multi-Atlas Segmentation}
\label{sec:multiatlas}

In this section, we first revisit the labelling problem in existing multi-atlas segmentation methods \cite{Heckemann2006multiatlas,Aljabar2009atlasselection,Artaechevarria2009,VanderLijn2008,Lotjonen2010} and reformulate it as an MRF energy optimisation problem defined on a graph comprising multiple images (i.e. the target and atlases). In particular, we show how the proposed graphical approach can incorporate label fusion (Sec.~\ref{sec:labelfusion}), spatial regularisation (Sec.~\ref{sec:regularisation}), as well as a data term and missing atlas labels (Sec.~\ref{sec:data-term-missing-labels}). Section~\ref{sec:framework-summary} summarises the components of the proposed framework. To solve the optimisation problem, in Sec.~\ref{sec:optimisation} we propose an extension of CMF~\cite{Yuan2010a} which can efficiently minimise energy functions on graphs connecting multiple images.

%%%%%%%%%%%%%%%%%%%%%%%%%%%%%%%%%%%%%%%%%%%%%%%%%%%%%%%%%%%%%%%%%%%%%%%%%%%%%%%%%%%%%%%%%%%%%%%%%%%
% *** Label Fusion ***
%%%%%%%%%%%%%%%%%%%%%%%%%%%%%%%%%%%%%%%%%%%%%%%%%%%%%%%%%%%%%%%%%%%%%%%%%%%%%%%%%%%%%%%%%%%%%%%%%%%
\subsection{Label Fusion}
\label{sec:labelfusion}

\begin{figure}[t]
 \centering
  % \setbox9=\hbox{\includegraphics[width=.3\linewidth]{example-image-1x1}}% Capture tallest image in box 9
  \setbox9=\hbox{\includegraphics[width=.3\linewidth]{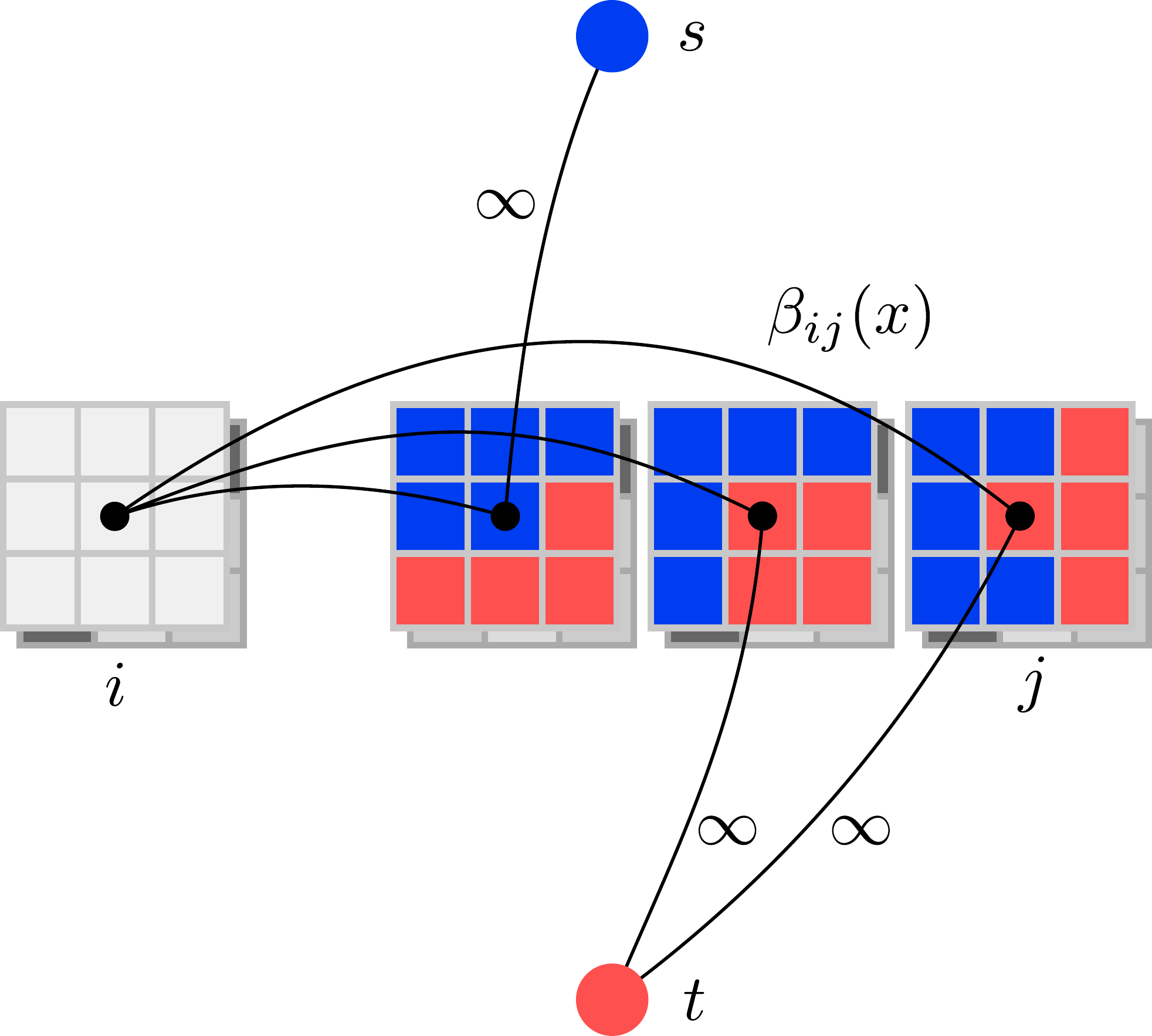}}% Capture tallest image in box 9
  \subcaptionbox{Images and label maps\label{fig:dataset}}
  {\raisebox{11mm}{\includegraphics[width=.40\textwidth]{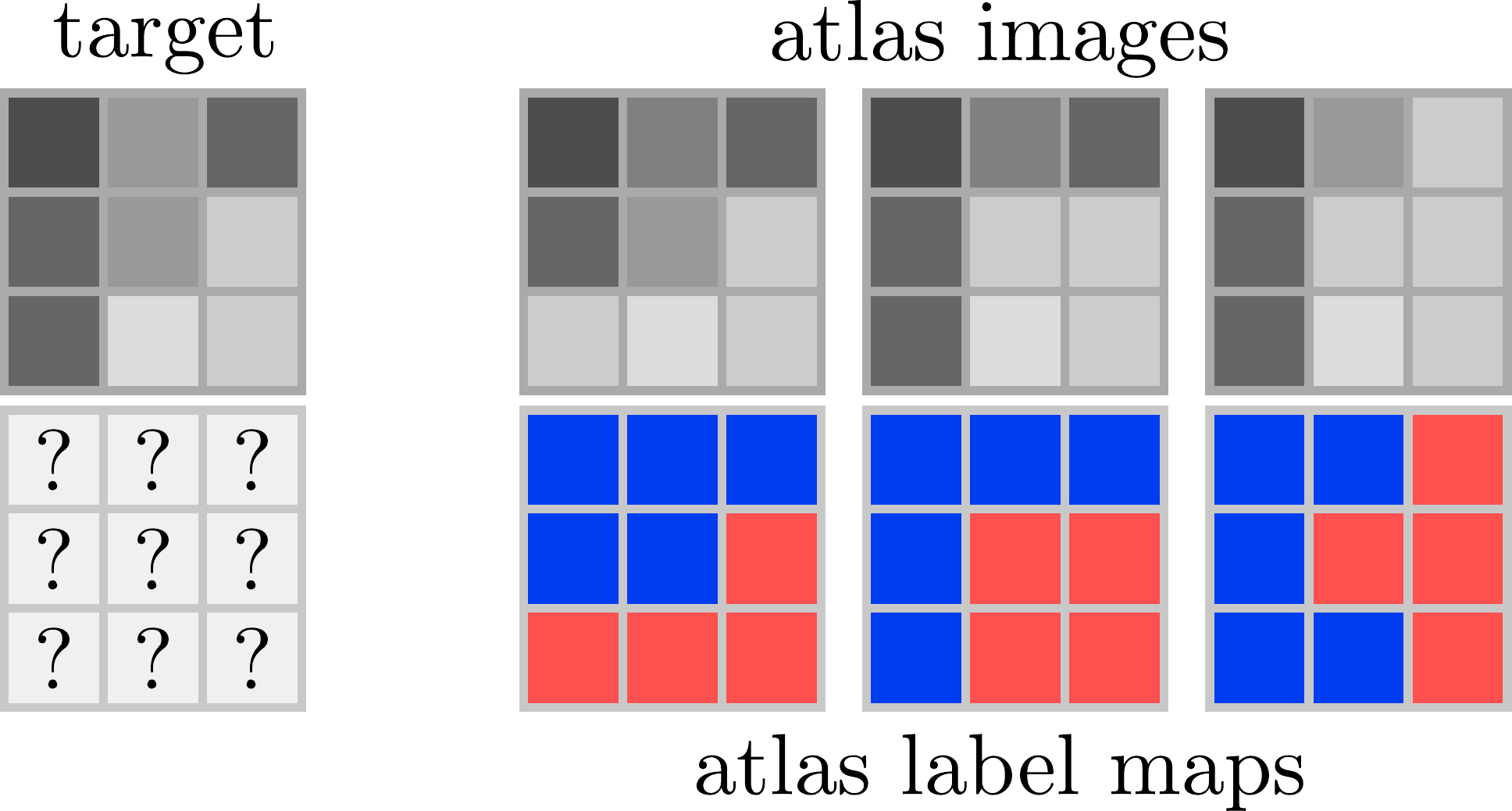}}} %\hfill
  \quad%
  \subcaptionbox{Graphical representation of label fusion\label{fig:multiatlas}}
  {\includegraphics[width=.40\textwidth]{figures/20141218_multiatlas}} %\hfill
  \caption{(a) A toy dataset with an unlabelled target image on the left,
    atlas images and corresponding manual annotations 
    (blue and red depict different labels) 
    on the
    right. (b) In MAS, each voxel $x$ in target image $i$ is labelled
    by label propagation from atlases $j \in \{1, \dotsc,R\}$ with
    fusion weights $\beta_{ij}(x)$. This can also be interpreted as an MRF optimisation problem, where atlas voxels are connected to the
    terminal nodes with infinitely weighted edges and inter-image edges $\beta_{ij}(x)$ encode label fusion.}
  \label{fig:methods}
\end{figure}

For multi-atlas segmentation \cite{Heckemann2006multiatlas,Artaechevarria2009} (MAS) using $R$ images, all atlas images $j \in
\{1, \dotsc, R\}$ are registered to the target image $i$. For convenience we assume $i=R+1$. The
label maps $l_j$ associated with the atlas images $j$ are then
propagated to the target. Figure\,\ref{fig:dataset} shows an example
atlas set with corresponding label maps, and an unlabelled target
image. Each voxel $x \in \Omega$ in the target image $i$ is labelled
using some combination strategy, e.g. a weighted average of atlas
labels $l_j(x)$:
\begin{align}
l_i(x) &= \argmax_L \sum_{j=1}^R \beta_{ij}(x) \delta( l_j(x) = L )
\label{eq:labelfusion}
\end{align}
Here $\delta(.)$ is an indicator function. The weights $\beta_{ij}(x)$ can be uniform (which is equivalent to the majority vote rule as used in \cite{Heckemann2006multiatlas,Aljabar2009atlasselection,Rohlfing2004}) or based on global or local similarity measures between images $i$ and $j$ as in \cite{Artaechevarria2009,Sabuncu2010,Bai2013}.

\begin{figure}[t]
 \centering
 {\includegraphics[width=.29\textwidth]{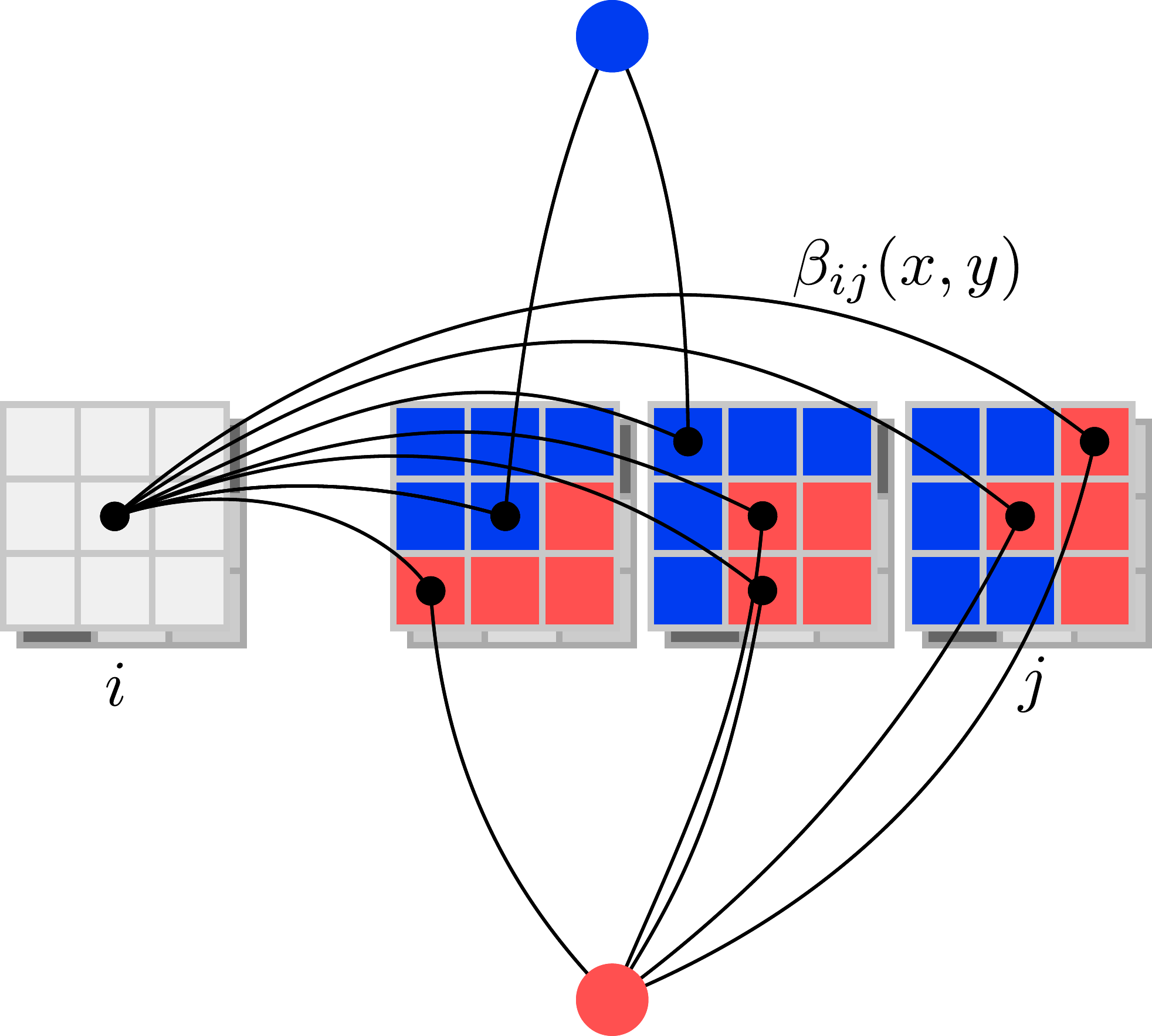}} %\hfill
 \caption{Graph configuration representing patch-based segmentation. $\beta_{ij}(x,y)$ is determined by a patch similarity measure between a patch centred around voxel $x$ in image $i$ and voxel $y$ in image $j$. Not all connections are drawn for better visibility and to reflect the fact that in practice, dissimilar patches are omitted in the label fusion~\cite{Coupe2011PatchBased}.}
 \label{fig:pbs-beta}
\end{figure}

As an alternative perspective, we can use a graphical representation to model the relationship of shared information between the atlases and the target using a Markov Random Field~\cite{Li1994}. 
According to the above labelling scenario, this graph connects each voxel $x$ in the target image $i$ to the corresponding voxels in the atlas images $j$ with an edge weighted by
$\beta_{ij}(x)$.
The manual annotations in the atlases can be encoded by the unary potential function
\begin{align}
  V ( l_j(x) ) &= \left\{
  \begin{array}{l l}
  0 & \quad l_j(x) = G_j(x) ,\\
  \infty & \quad \mbox{otherwise}
  \label{eq:unary_potential}
  \end{array} \right.
\end{align}
where $G_j(x)$ is the ground truth label given by the expert rater, assigning infinite cost to the hypothetical scenario of assigning a different label to the atlas voxel.
Figure\,\ref{fig:multiatlas} visualises this configuration and in Sec.~\ref{sec:data-term-missing-labels}, these terminal graph connections are discussed in more detail. To find a labelling on the graph, we can formulate a pairwise potential function that penalises conflicting labels in voxels connected by a high weight $\beta_{ij}(x)$, e.g.
\begin{equation}
V (l_i(x), l_j(x) ) = \beta_{ij}(x) \delta( l_j(x) \neq l_i(x) )
\label{eq:pairwise_potential}
\end{equation}
This assigns a high penalty when the target and atlas labels differ and the atlas is considered similar to the target $i$, as defined by the similarity measure $\beta_{ij}(x)$.
In the case of a majority vote, the weights are uniform,
e.g. $\beta_{ij}(x)=1$.  The cost for labelling an individual voxel
$x$ in image $i$ can then be calculated as follows:
\begin{align}
E_{\text{propagation}} \left( l_i(x) \right) &= 
  \sum_{j=1}^R V (l_i(x), l_j(x) ) \\
  &= \sum_{j=1}^R \beta_{ij}(x) \delta( l_j(x) \neq l_i(x) ) \\
  &= \sum_{j=1}^R \beta_{ij}(x) - \sum_{j=1}^R \beta_{ij}(x) \delta( l_j(x) = l_i(x) )
\end{align}
As we assume the graph satisfies Markov properties, voxels in the target image are conditionally independent given the atlas images since spatially neighbouring voxels in the target image are not connected in the graph (in contrast to the setting for regularisation in many vision problems\,\cite{Li1994}). 
Since the atlas labels are fixed and assumed to be independent of each other (a common assumption in MAS), it follows that the target voxels are statistically independent, and the optimal label can be found by
minimising $E_{\text{propagation}} \left( l_i(x) \right)$
independently for all voxels:
\begin{align}
l_i(x) &= \argmin_L \quad E_{\text{propagation}} \left( l_i(x) = L \right) \\
&= \argmin_L \quad - \sum_{j=1}^R \beta_{ij}(x) \delta( l_j(x) = L ) \\
&= \argmax_L \quad \sum_{j=1}^R \beta_{ij}(x) \delta( l_j(x) = L ) ~. 
\label{eq:labelfusion_proof}
\end{align}
This leads to the same result as the vote rule in Eq.\,\ref{eq:labelfusion}, demonstrating that multi-atlas segmentation can be expressed in terms of a graph optimisation problem. 
It is important to note that patch-based segmentation\,(PBS\,\cite{Coupe2011PatchBased,Rousseau2011}) can also be expressed in this framework. In this case we use a slightly different graph structure as the label fusion step in PBS takes into account multiple voxels in a neighbourhood of $x$ in each atlas instead of just one voxel at location $x$. By denoting the patch-based label fusion weights as $\beta_{ij}(x,y), y \in \mathcal{N}_x$ to reflect the non-local nature of these methods, a labelling can be found for this scenario as well. Here, multiple patches in the atlases are used at locations $y$ in a neighbourhood $\mathcal{N}_x$ around location $x$. This scenario is visualised in Fig.~\ref{fig:pbs-beta}. While the proposed formulation holds for these non-local techniques, the graph structure becomes more complex. In the scope of this paper, we limit ourselves to graphs on regular grids where voxels in different images are only connected if they are at corresponding locations, as this makes it possible to use the efficient optimisation scheme proposed in Sec.~\ref{sec:optimisation}.

This novel perspective on label fusion for multi-atlas segmentation has two advantages: (1) it allows easy integration of additional components and therefore provides a unifying reformulation for existing multi-atlas segmentation methods, and (2) the graphical approach extends to segmentation using partially annotated atlases (Sec.\,\ref{sec:segmentation_using_partially_annotated_atlas_data}).

%%%%%%%%%%%%%%%%%%%%%%%%%%%%%%%%%%%%%%%%%%%%%%%%%%%%%%%%%%%%%%%%%%%%%%%%%%%%%%%%%%%%%%%%%%%%%%%%%%%
% *** Spatial Regularisation ***
%%%%%%%%%%%%%%%%%%%%%%%%%%%%%%%%%%%%%%%%%%%%%%%%%%%%%%%%%%%%%%%%%%%%%%%%%%%%%%%%%%%%%%%%%%%%%%%%%%%
\subsection{Spatial Regularisation}
\label{sec:regularisation}

\begin{figure}[t]
 \centering
  \subcaptionbox{Multi-atlas segmentation with regularisation\label{fig:multiatlas-reg}}
  {\includegraphics[width=.29\textwidth]{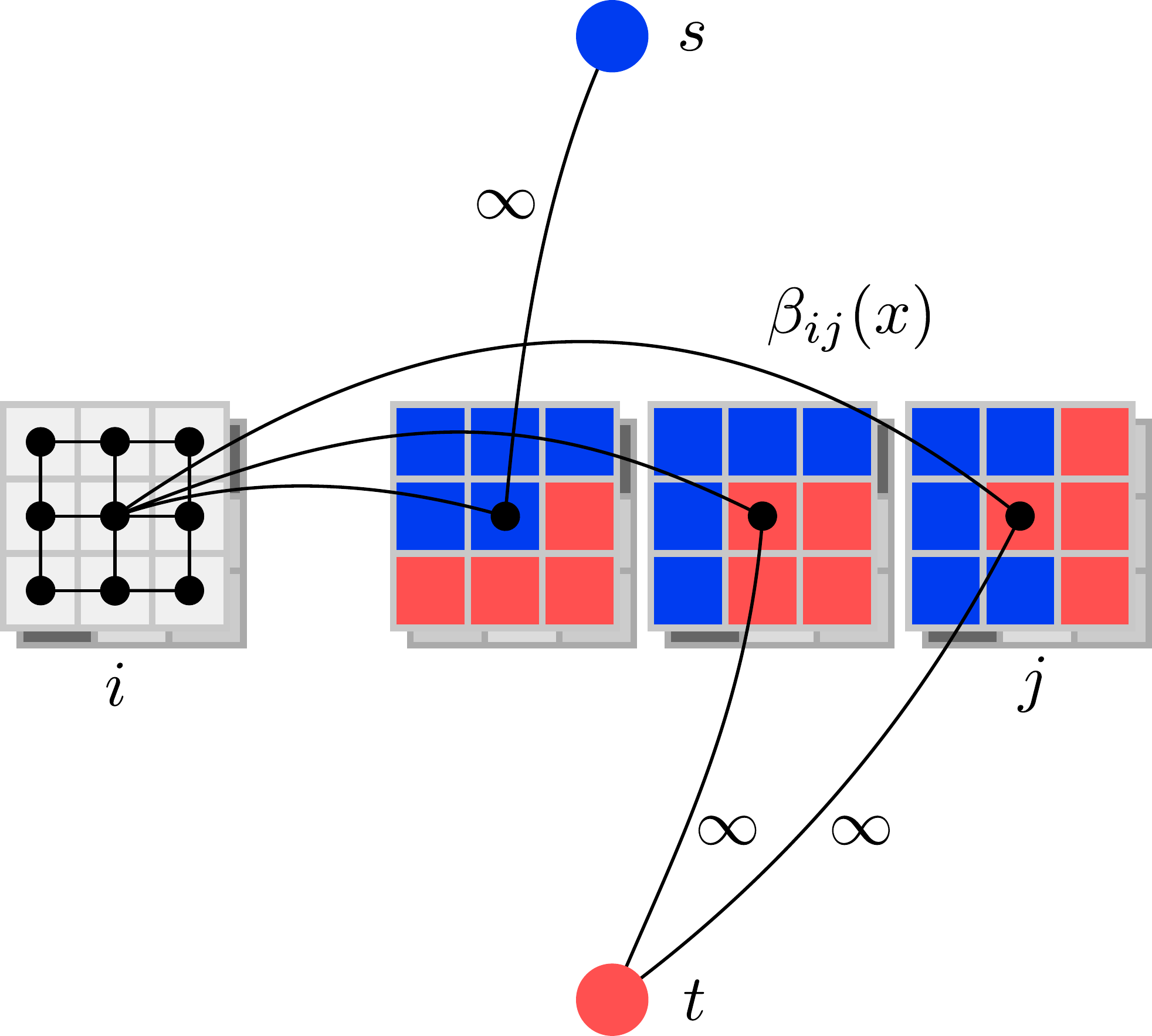}} %\hfill
  \quad%
  \subcaptionbox{Additional data term\label{fig:multiatlas-data}}
  {\includegraphics[width=.29\textwidth]{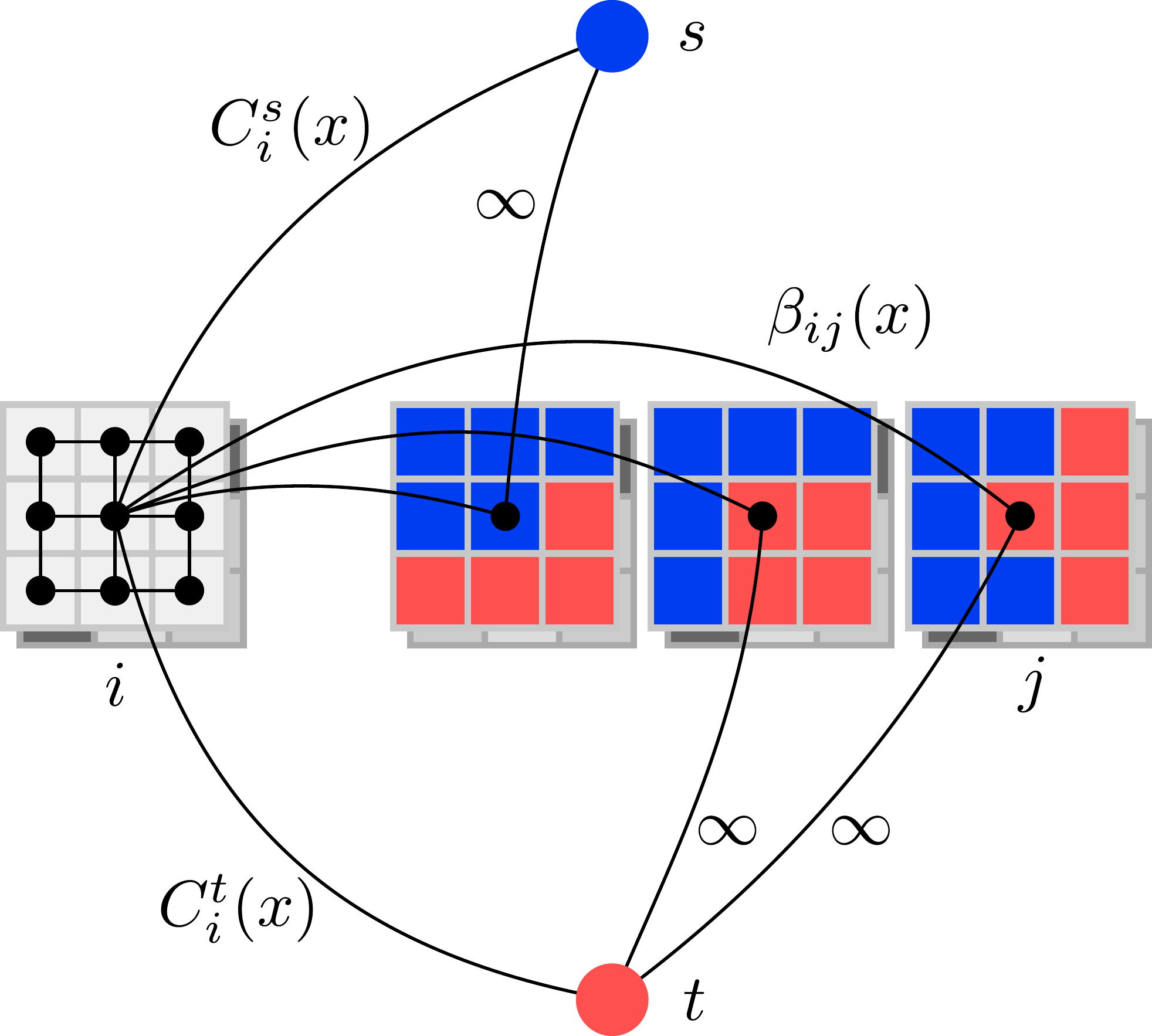}} %\hfill
  \quad%
  \subcaptionbox{Missing labels\label{fig:multiatlas-missing-labels}}
  {\includegraphics[width=.29\textwidth]{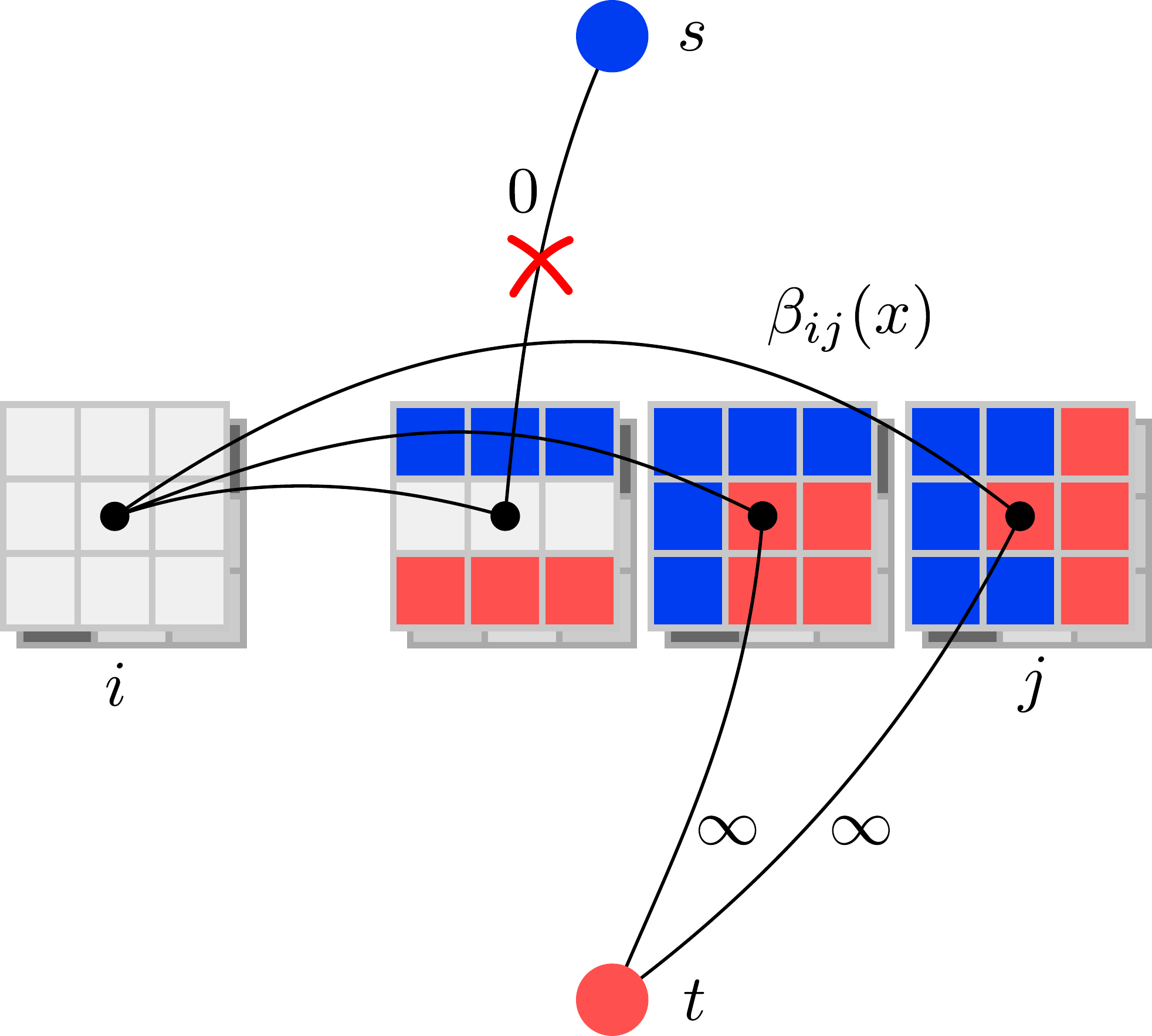}} %\hfill
  \quad%
  \caption{This figure shows different graph configurations representing (a) multi-atlas segmentation with spatial regularisation in the target image, (b) an additional data term in the target image, i.e. encoding intensity models for the data, (c) multi-atlas segmentation with missing atlas labels. Missing labels are reflected in the graph structure by missing terminal connections.}
  \label{fig:multiatlas-reg-data}
\end{figure}

In the previous section, we proposed assigning pairwise potentials between target and atlas voxels
for label propagation. In addition, we can incorporate spatial regularisation with pairwise
potentials between adjacent voxels within an image. This simple modification of the graph structure
is shown in Fig.~\ref{fig:multiatlas-reg}.
Regularisation enforces spatial consistency by penalising different label assignment in adjacent voxels. If the regularisation weights are based on intensity gradients, consistent labels can be enforced in adjacent labels that are similar in appearance, while allowing different labels across intensity boundaries.
A graph configuration as shown in Fig.~\ref{fig:multiatlas-reg} models the scenario where regularisation is used to refine label fusion results, as for example in \cite{VanderLijn2008,Wolz2013,Wang2014}.

%%%%%%%%%%%%%%%%%%%%%%%%%%%%%%%%%%%%%%%%%%%%%%%%%%%%%%%%%%%%%%%%%%%%%%%%%%%%%%%%%%%%%%%%%%%%%%%%%%%
% *** Data Term and Missing Labels ***
%%%%%%%%%%%%%%%%%%%%%%%%%%%%%%%%%%%%%%%%%%%%%%%%%%%%%%%%%%%%%%%%%%%%%%%%%%%%%%%%%%%%%%%%%%%%%%%%%%%
\subsection{Data Term and Missing Labels}
\label{sec:data-term-missing-labels}

In Eq.~\ref{eq:unary_potential} we showed how manual annotations can be encoded as unary potentials which are often referred to as a data term~\cite{Li1994,Boykov2000}. The ground truth nature of these annotations is reflected in the graph structure by infinitely weighted terminal connections for each atlas voxel according to the manual label given. As can be seen in Fig.~\ref{fig:multiatlas} or \ref{fig:multiatlas-reg}, the voxels in the target image are not connected to the terminals as they are assumed to be unlabelled and no prior knowledge is available for them. It is important to note that a data term can be specified for the target image as well using prior probabilities, intensity models of the data, or a combination of both. This is a common technique when using MRFs in vision problems~\cite{Li1994,Kolmogorov2004,VanderLijn2008,Lotjonen2010} and can be incorporated by extending the graph structure as visualised in Fig.~\ref{fig:multiatlas-data}. Furthermore, missing labels can be easily accounted for by removing terminal connections (i.e. unary potentials) for voxels where annotations are not available, as shown in Fig.~\ref{fig:multiatlas-missing-labels}. The important implications of this property will be discussed in detail in Sec.~\ref{sec:segmentation_using_partially_annotated_atlas_data} in conjunction with partially annotated atlas data.

%%%%%%%%%%%%%%%%%%%%%%%%%%%%%%%%%%%%%%%%%%%%%%%%%%%%%%%%%%%%%%%%%%%%%%%%%%%%%%%%%%%%%%%%%%%%%%%%%%%
% *** Summary ***
%%%%%%%%%%%%%%%%%%%%%%%%%%%%%%%%%%%%%%%%%%%%%%%%%%%%%%%%%%%%%%%%%%%%%%%%%%%%%%%%%%%%%%%%%%%%%%%%%%%
\subsection{Summary} % (fold)
\label{sec:framework-summary}

We propose to interpret both the target image and the set of atlas images as a single graph structure (in which each voxel is a node) satisfying Markov properties. On this graph we can use unary potentials to define the data term $E_{\text{data}}$ to encode manual annotations or other prior knowledge, or to reflect missing labels. We then showed how pairwise potentials can be used to encode label fusion through \emph{inter-image} connections and to build a propagation energy term $E_{\text{propagation}}$. Another pairwise potential term $E_{\text{regularisation}}$ encodes spatial regularisation through \emph{intra-image} edges. The propagation, data, and regularisation terms can be combined to a comprehensive labelling energy function defined for the whole graph:
\begin{equation}
E( l )  = E_{\text{data}}(l) + E_{\text{regularisation}}(l) +  E_{\text{propagation}}(l) 
\label{eq:unified_framework}
\end{equation}
As mentioned in the introduction, many existing multi-atlas segmentation methods (e.g.\,\cite{Makropoulos2014,VanderLijn2008}) use an MRF formulation to improve label propagation results with the benefits of regularisation and intensity data models. However, these approaches use probabilistic label propagation results as prior probabilities (i.e. unary potentials) in a \emph{subsequent} refinement step, therefore adding the MRF optimisation as a separate post-processing step. The above comprehensive formulation treats label propagation as part of the optimisation process, and unifies all the components within a single framework. Furthermore, as we show in Sec.~\ref{sec:segmentation_using_partially_annotated_atlas_data}, the flexibility of the proposed graph structure lends itself naturally to exploit partially annotated data.

%%%%%%%%%%%%%%%%%%%%%%%%%%%%%%%%%%%%%%%%%%%%%%%%%%%%%%%%%%%%%%%%%%%%%%%%%%%%%%%%%%%%%%%%%%%%%%%%%%%
% *** Optimisation using Continuous Max-Flow (CMF) ***
%%%%%%%%%%%%%%%%%%%%%%%%%%%%%%%%%%%%%%%%%%%%%%%%%%%%%%%%%%%%%%%%%%%%%%%%%%%%%%%%%%%%%%%%%%%%%%%%%%%
\subsection{Optimisation using Continuous Max-Flow (CMF)}
\label{sec:optimisation}

\begin{figure}[t]
 \centering
 {\includegraphics[height=.15\textheight]{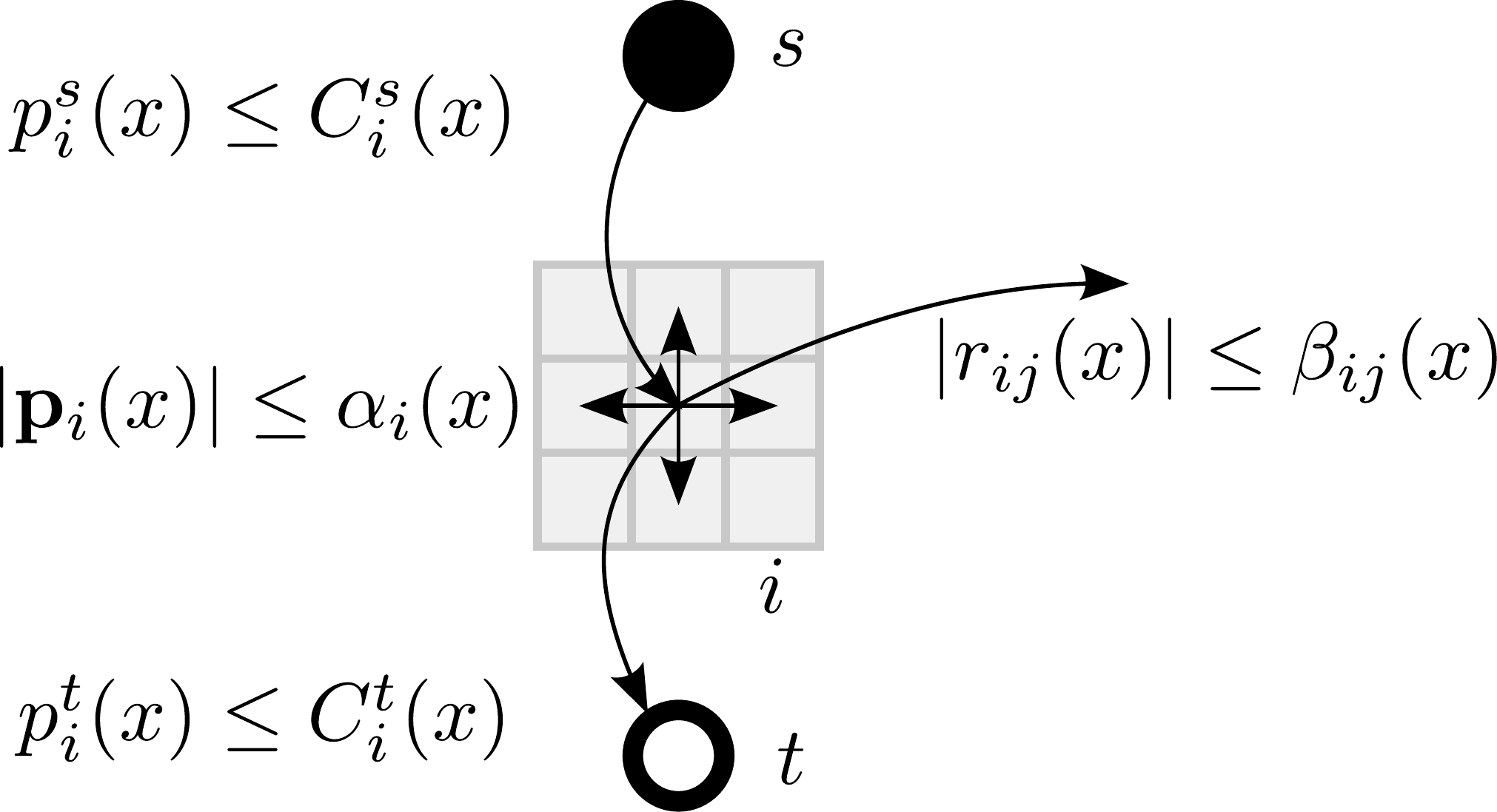}} %\hfill
 \caption{Notation for flow constraints $\beta_{ij}(x), C_i^{s,t}(x), \alpha_i(x)$  for label propagation, data term and spatial regularisation,  and corresponding inter-image flows $r_{ij}(x)$, source and sink flows $p_i^{s,t}(x)$ and spatial flows $\mathbf{p}_i(x)$, respectively, at location
   $x$ in image $i$.}
 \label{fig:notation_node}
\end{figure}

It has been shown that MRF energy functions consisting of unary and pairwise terms can be minimised using min-cut/max-flow approaches if the pairwise terms are metric or semi-metric\,\cite{Boykov2001}, yielding globally optimal results for binary labelling problems and approximately globally optimal results for multiple labels\,\cite{Boykov2001}. Recently, \cite{Yuan2010} proposed a continuous max-flow (CMF) algorithm in the 2D or 3D domain (i.e. a single image) which avoids metrication bias and is inherently parallelisable in contrast to many discrete graph-based methods\,\cite{Yuan2010}. As the proposed energy function needs to be optimised for a large graph consisting of voxels in all images and their interactions, this approach was adopted and extended for graphs between multiple images.

Analogous to discrete max-flow approaches, the energy function on the graph can be optimised by
maximising a source flow $p^s$ through the network, subject to flow conservation and capacity
constraints on the edges. In the original CMF algorithm\,\cite{Yuan2010}, spatial flows $\mathbf{p}
= [ p_x, p_y, p_z ]^T$ exist between adjacent voxels in the image domain $\Omega$ (for
regularisation) and source and sink flows $p^{s,t}$ between voxels and terminal nodes. The
optimisation is performed with a variational approach by introducing a Lagrange multiplier $u(x)$ to
incorporate the constraints\,\cite{Yuan2010}. It has been shown that the resulting $u(x)$
corresponds to the globally optimal labelling\,\cite{Yuan2010} in the binary case.

\subsubsection{Binary segmentation using CMF}

In the following, we propose a generalisation of CMF from a single image to an arbitrary
configuration of interconnected images to account for any user-defined choice of inter-image
relationships $\beta_{ij}(x)$.  Figure\,\ref{fig:notation_node} shows the capacity constraints and
introduces the notation for inter-image flows $r_{ij}(x)$ (for label propagation), spatial flows
$\mathbf{p}_i(x)$ (for regularisation) and terminal flows $p_i^{s,t}(x)$ (for the data term). The
notation is similar to~\cite{Trus2014}, where inter-image constraints were used in a different
context. To satisfy flow conservation, the sum of all in- and outgoing flows $\rho_i(x)$ at each
node must be zero, i.e.
\begin{align}
\rho_i(x) = 
\divg \mathbf{p}_i(x) - p_i^s(x) + p_i^t(x) + \sum_{j=1,j\neq i}^{n} r_{ij}(x) = 0,
\end{align}
where $r_{ij}(x) = -r_{ji}(x)$ and $n$ is the number of images in the graph. We propose to adapt the
definitions of the discrete gradient and divergence operators to account for anisotropic voxel
dimensions $[s_x,s_y,s_z]$, which are often found in medical images:
\begin{align}
\nabla \mathbf{p} &= \left[ \frac{\delta_x \mathbf{p}}{s_x}, \frac{\delta_y \mathbf{p}}{s_y} , \frac{\delta_z \mathbf{p}}{s_z} \right]^T \\
\divg \mathbf{p} &= \nabla \cdot \mathbf{p}
\end{align}
This leads to the Lagrangian function
\begin{align}
L(u, p^s,p^t,\mathbf{p},r) &= 
\sum_{i=1}^{n} \left( \int_{\Omega} p_i^s dx + <u_i,\rho_i> - \frac{c}{2} \| \rho_i \|^2    \right),
\label{eq:lagrangian}
\end{align}
which can be maximised iteratively by optimising each variable $u, p^s,p^t,\mathbf{p},r$
separately\,\cite{Trus2014,Yuan2010}. The spatial flows $\mathbf{p}_i(x)$ are updated using the
gradient projection approach proposed in \cite{Chambolle2004}:
\begin{align}
\mathbf{p}^{k+1} = \argmax_{\|p(x)\| \leq \alpha(x)} - \frac{c}{2} \| \divg \mathbf{p}^k - F^k \|^2
\end{align}
The regularisation constraints $\alpha(x)$ determine the smoothness of the result. To enforce greater smoothness in homogeneous image regions than along intensity boundaries, $\alpha(x)$
can be defined based on the image gradient $\nabla I(x)$:
\begin{align}
\alpha(x) &= a \exp \left( - \frac{ \| \nabla I(x) \|^2}{2 \sigma_1^2}  \right)
\label{eq:exp_alpha}
\end{align}
with parameters $a$ and $\sigma_1$. This measure is the continuous equivalent of the
regularisation term used in in~\cite{VanderLijn2008}, one of the pioneering works combining
regularisation and multi-atlas segmentation. The terminal flows $p_i^s(x), p_i^t(x)$ can also be
found by fixing all other variables, respectively \cite{Yuan2010}. The novel component compared
to\,\cite{Trus2014,Yuan2010} is the use of inter-image flows $r_{ij}(x)$ between any pair of images
$i,j$~\cite{Koch2015}. We therefore show in particular that the optimisation step at iteration $k$ for $r_{ij}(x)$, while fixing all other variables, is:
\begin{alignat}{3}
&r_{ij}^{k+1} &&=  \argmax_{|r_{ij}| \le \beta_{ij} } ~&& L(u, p^s,p^t,\mathbf{p},r)
\end{alignat}
This leads to 
\begin{align}
r_{ij}^{k+1} =&  \left\{
    \begin{array}{l l}
    - \beta_{ij}, & \quad \frac{1}{2}(J_j^k-J_i^k) \le -\beta_{ij} ~ ,\\
    \frac{1}{2}(J_j^k-J_i^k), & \quad | \frac{1}{2}(J_j^k-J_i^k) | \le \beta_{ij} ~ ,\\
    \beta_{ij} & \quad \mbox{otherwise.}
    \end{array} \right.
\end{align}
where 
\begin{align}
J_i^k = 
(\divg \mathbf{p}_i - p_i^s + p_i^t)^k + \sum_{l=1,l\neq i,j}^{n} r_{il}^{k} - \frac{u_i^k}{c}
\end{align}
%
% A more detailed derivation of this result is given in the supplemental material. 
The multiplier $u_i(x)$, which serves as the labelling function, is updated with
\begin{align}
u_i(x)^{k+1} = u_i(x)^{k} - c \rho_i(x)^k
\end{align}
After convergence, a segmentation can be found by discretising the resulting solution for
$u$, e.g. by thresholding at $50\%$.

\subsubsection{Multi-label segmentation using the Potts Model}

\begin{figure}[ht]
 \centering
 {\includegraphics[height=.15\textheight]{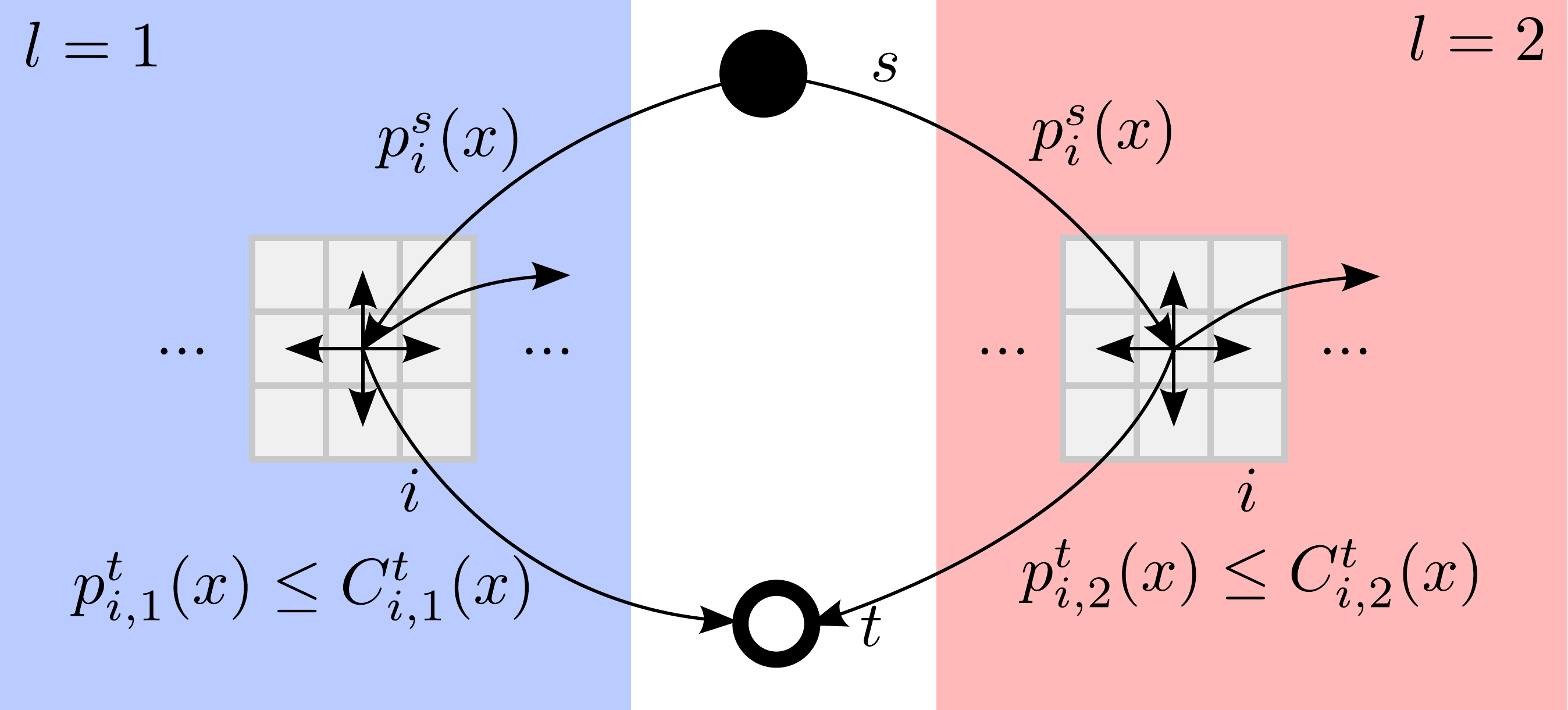}} %\hfill
 \caption{Schematic showing graph configuration for multi-label CMF using the Pott's Model. The graph (in this figure only one image $i$ is shown) is replicated for each label $l$. The data term is encoded in the sink constraints for every label.}
\label{fig:multi-label-cmf}
\end{figure}

CMF has been extended to multi-label segmentation problems in~\cite{Yuan2010a} using a Pott's model
approach. To optimise for multiple labels, the graph structure is duplicated for every label. The
data term is encoded in the sink constraints of each ``sub-graph'' while the source connections
remain unconstrained. The same changes can be applied to the the graph in our framework, as shown in
Fig.~\ref{fig:multi-label-cmf}.

The Lagrangian function formulated for the binary case (Eq.~\ref{eq:lagrangian}) can be augmented to
reflect this graph configuration:
\begin{align}
\begin{split}
L(u, p^s,p^t,\mathbf{p},r) =& 
\sum_{i=1}^{n} \left( \int_{\Omega} p_i^s dx + \sum_{l=1}^{L} <u_{i,l},\rho_{i,l}>  - \frac{c}{2}  \sum_{l=1}^{L}  \| \rho_{i,l} \|^2    \right) ~.
\end{split}
\label{eq:lagrangian-ml}
\end{align}
Here, $u_{i,l}$ is the labelling function for label $l \in {1, .., L}$ in image $i$ and $\rho_{i,l}$
is the new flow conservation constraint
\begin{align}
\rho_{i,l}(x) = 
\divg \mathbf{p}_{i,l}(x) - p_i^s(x) + p_{i,l}^t(x) + \sum_{j=1,j\neq i}^{n} r_{ij,l}(x) = 0 ~.
\end{align}

%%%%%%%%%%%%%%%%%%%%%%%%%%%%%%%%%%%%%%%%%%%%%%%%%%%%%%%%%%%%%%%%%%%%%%%%%%%%%%%%%%%%%%%%%%%%%%%%%%%
%%%%%%%%%%%%%%%%%%%%%%%%%%%%%%%%%%%%%%%%%%%%%%%%%%%%%%%%%%%%%%%%%%%%%%%%%%%%%%%%%%%%%%%%%%%%%%%%%%%
% *** PARTIAL ANNOTATION STRATEGIES ***
%%%%%%%%%%%%%%%%%%%%%%%%%%%%%%%%%%%%%%%%%%%%%%%%%%%%%%%%%%%%%%%%%%%%%%%%%%%%%%%%%%%%%%%%%%%%%%%%%%%
%%%%%%%%%%%%%%%%%%%%%%%%%%%%%%%%%%%%%%%%%%%%%%%%%%%%%%%%%%%%%%%%%%%%%%%%%%%%%%%%%%%%%%%%%%%%%%%%%%%

\section{Partial Annotation Strategies}
\label{sec:segmentation_using_partially_annotated_atlas_data}

Manually annotating medical images is very time consuming, placing a major burden on clinical
experts tasked with labelling large datasets. However, using the proposed unified framework for
multi-atlas segmentation, it is possible to open up a new field of applications, namely segmentation
using partially annotated atlas data. We showed in Sec.~\ref{sec:data-term-missing-labels} how the
proposed graphical representation can easily accommodate missing labels through missing terminal
connections in the graph structure. By applying our framework to any of the existing approaches
discussed throughout Sec.~\ref{sec:multiatlas}, this would lead to a segmentation that is inferred from the \emph{available} labels only, ignoring missing information.

\begin{figure}[t]
 \centering
 \subcaptionbox{Graph configuration 1 (CONF1)\label{fig:partial_config_a}}
 {\includegraphics[height=.24\textheight]{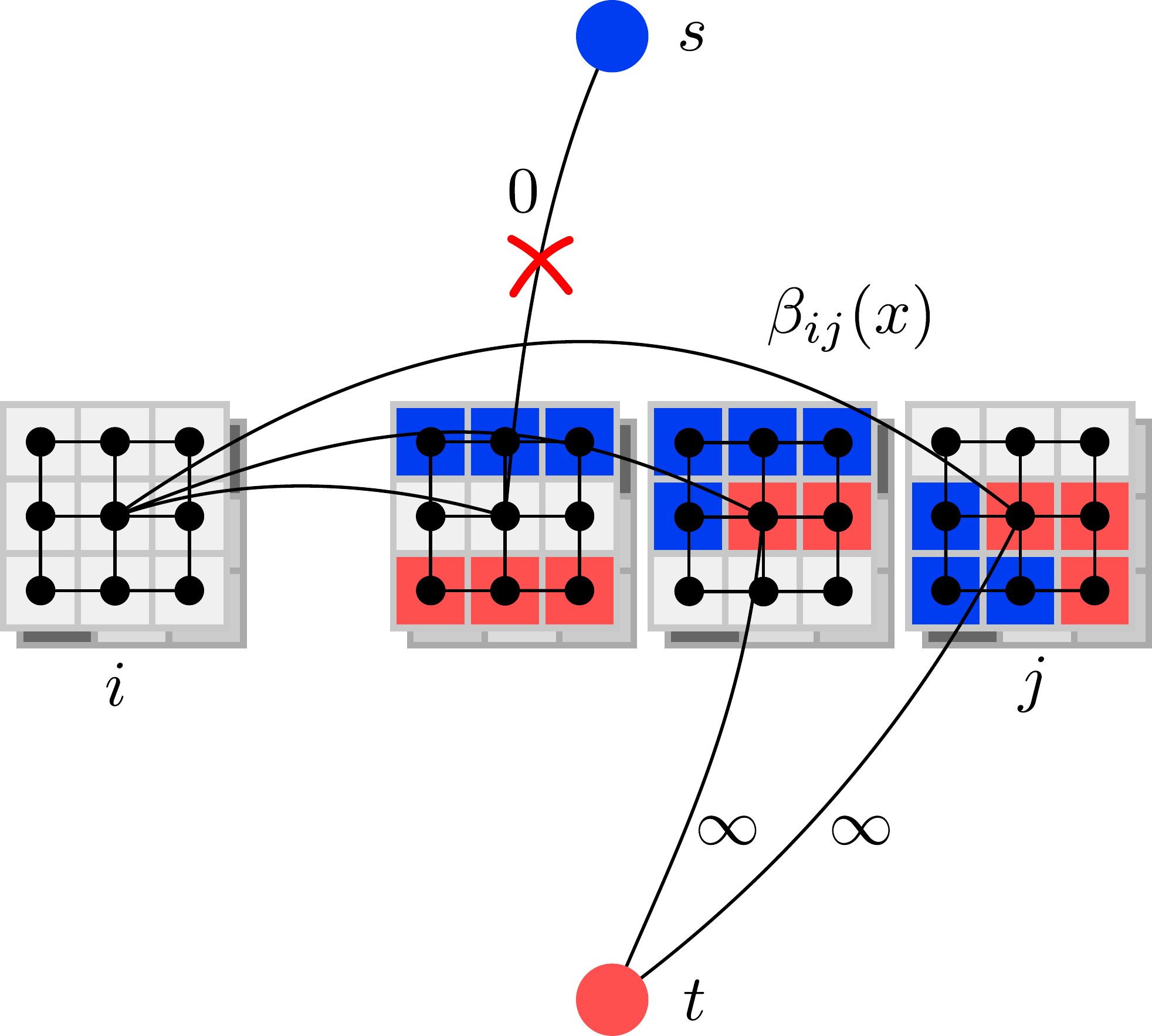}} %\hfill
 \quad\quad\quad%
 \subcaptionbox{Graph configuration 2 (CONF2)\label{fig:partial_config_b}}
 {\includegraphics[height=.24\textheight]{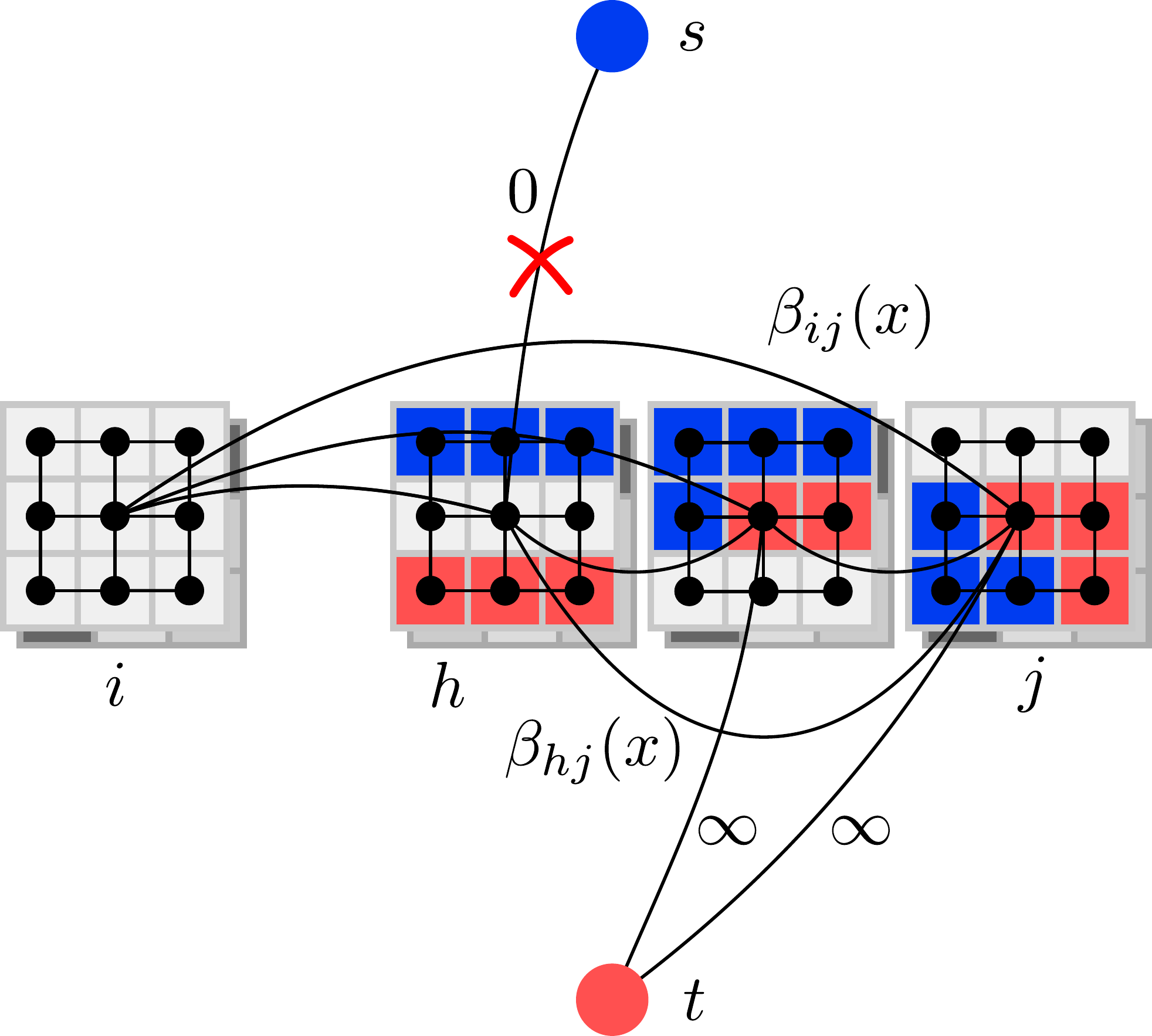}} %\hfill
 \caption{This figure shows two graph configurations used when employing partially annotated atlas data (blue and red depict different labels), based on the example dataset of Fig.\,\ref{fig:dataset}. Voxels with missing labels (white) are disconnected from terminal
   nodes. In contrast to Fig.~\ref{fig:multiatlas-missing-labels}, spatial regularisation is enabled in \emph{all} images. (a) Voxels
   at each location $x$ in the target image are connected to voxels in
   atlases $j$. (b) Additionally, atlas voxels are connected to voxels
   in other atlases.}
  \label{fig:partial_configs}
\end{figure}

Additionally, spatial consistency in the atlas images can be exploited to employ unlabelled atlas
data as well. As neighbouring voxels are expected to share the same label, particularly if the
voxels exhibit similar intensity patterns, we propose to use spatial regularisation within the atlas
images as a form of \emph{intra-image} label propagation. This way, labels may be shared between
similar regions with labelled and unlabelled voxels in the atlases and propagated to the target
image. This modification in the graph structure leads to a configuration as shown in
Fig.~\ref{fig:partial_config_a}. Another possible configuration combines this with an additional
inter-atlas propagation scheme which allows atlases to share information as well (shown in
Fig.~\ref{fig:partial_config_b}). This serves to facilitate the propagation, especially when manual
labels are very scarce at some locations $x$.

With this framework, it becomes interesting to pursue strategies which aim to efficiently build partially annotated datasets which may then be used as training data for segmentation tasks. In the remainder of this section, we propose two partial annotation strategies, which are evaluated in the Experiments Sections~\ref{sec:exp-pa-slices} and \ref{sec:exp-pa-scribbles}.

\subsection{Strategy A: Slicewise Annotation}
\label{sec:strategy-slices}

Medical volumetric images are often manually annotated slice-by-slice. Therefore reducing the proportion of annotated slices while retaining robust and accurate segmentation is an important goal. To simulate partially annotated atlases, only annotations from a proportion of evenly spaced 2D slices are used, and the remaining labels are set to be ``missing''. As an example, Fig.~\ref{fig:annotation-strategy_a} shows a cross-section of a 3D image where every fifth slice is annotated. It is important to note that in the selected slices, the structures of interest are delineated in detail, i.e. all voxels in that slice are labelled.

\subsection{Strategy B: Scribbles}
\label{sec:strategy-scribbles}

\begin{figure}[t]
 \centering
 \subcaptionbox{Partial annotation strategy A: Slicewise\label{fig:annotation-strategy_a}}
 {\includegraphics[height=.17\textheight]{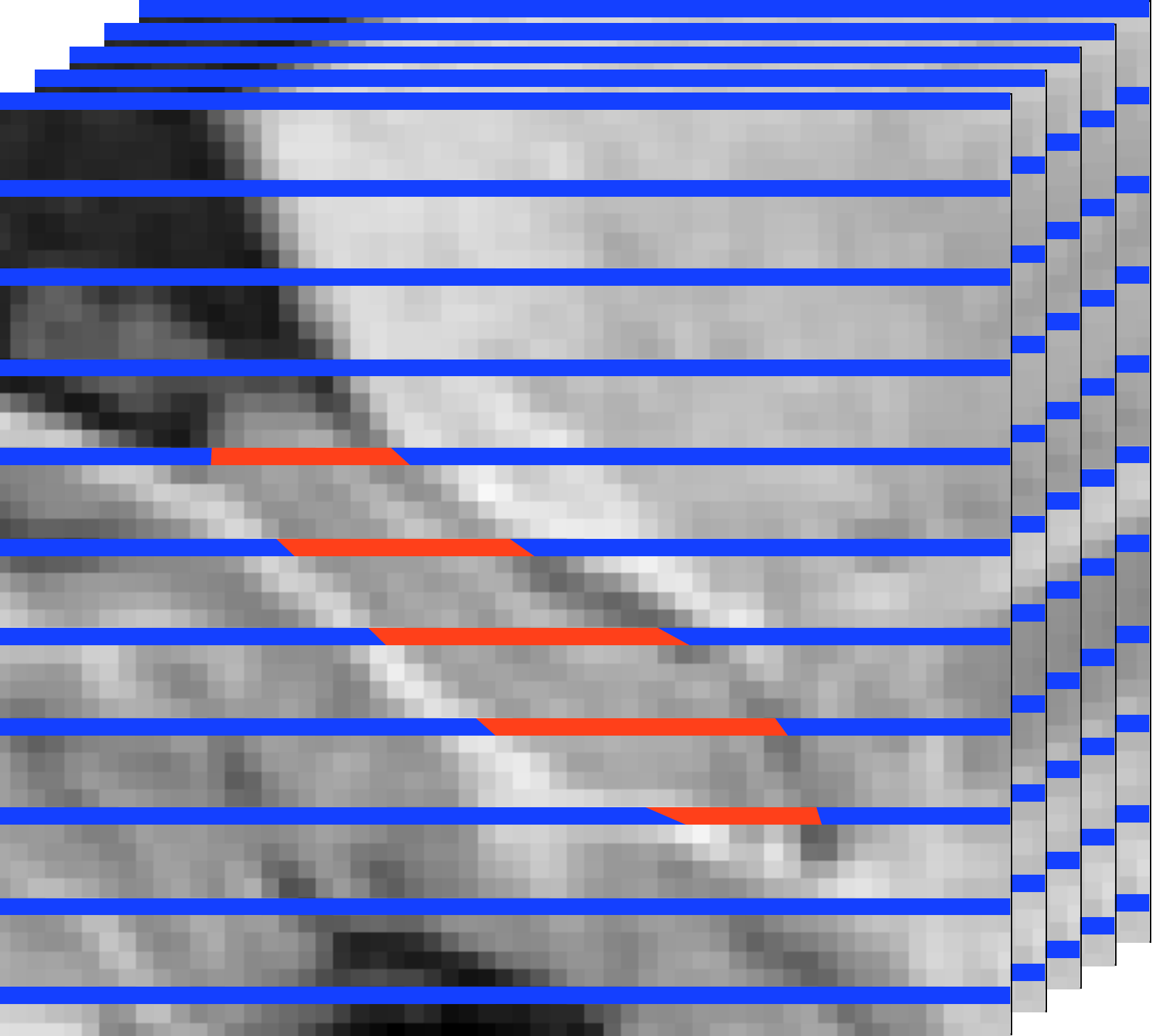}} %\hfill
 \quad\quad%
 \subcaptionbox{Partial annotation strategy B: Scribbles\label{fig:annotation-strategy_b}}
 {\includegraphics[height=.17\textheight]{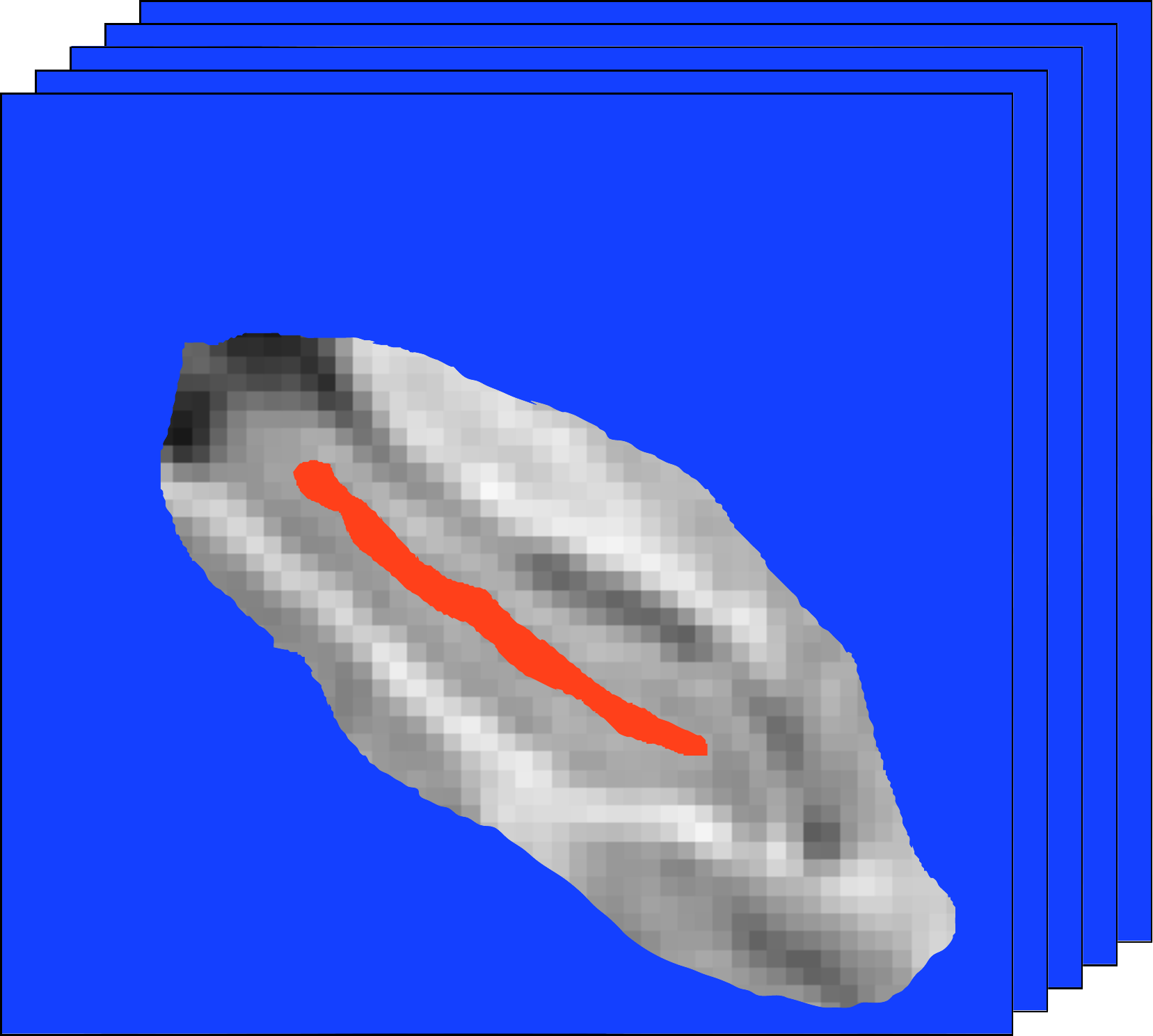}} %\hfill
 \caption{This figure visualises two partial annotation strategies: (a) shows a volumetric image with partial slice-by-slice annotation and (b) shows the same image with scribbles placed on each slice. Red and blue depict foreground and background, respectively, and voxels in grey remained unlabelled. }
  \label{fig:annotation-strategies}
\end{figure}

Scribbles are often used to annotate images in the context of interactive segmentation~\cite{Boykov2000,Rother2004}. This strategy typically involves placing brush strokes (i.e. ``scribbles'') on parts of the image considered within the structure of interest, or within the background. As scribbles do not delineate the structure boundary, this only requires a very short user interaction and could potentially require less expertise. These properties make ``scribbling'' an attractive annotation strategy if it can be shown their use leads to competitive segmentation results. Figure~\ref{fig:annotation-strategy_b} shows an example image with scribbles for both the structure of interest (i.e. the hippocampus) and the background.
We propose to annotate the training dataset by efficiently placing scribbles covering large areas (without delineating boundaries), as this can be done efficiently and is expected to make the segmentation task easier than very sparse, small scribbles.

%%%%%%%%%%%%%%%%%%%%%%%%%%%%%%%%%%%%%%%%%%%%%%%%%%%%%%%%%%%%%%%%%%%%%%%%%%%%%%%%%%%%%%%%%%%%%%%%%%%
%%%%%%%%%%%%%%%%%%%%%%%%%%%%%%%%%%%%%%%%%%%%%%%%%%%%%%%%%%%%%%%%%%%%%%%%%%%%%%%%%%%%%%%%%%%%%%%%%%%
% *** EXPERIMENTS AND RESULTS ***
%%%%%%%%%%%%%%%%%%%%%%%%%%%%%%%%%%%%%%%%%%%%%%%%%%%%%%%%%%%%%%%%%%%%%%%%%%%%%%%%%%%%%%%%%%%%%%%%%%%
%%%%%%%%%%%%%%%%%%%%%%%%%%%%%%%%%%%%%%%%%%%%%%%%%%%%%%%%%%%%%%%%%%%%%%%%%%%%%%%%%%%%%%%%%%%%%%%%%%%

\section{Experiments and Results}
\label{sec:experiment_and_results}

In the previous sections, we proposed a unified multi-atlas segmentation framework which can naturally accommodate partially annotated atlas data. We showed how the proposed graphical representation can implement a number of existing techniques through changes in the graph configuration. 
In the following experiments, we first employ the proposed framework to perform hippocampal segmentation using three existing multi-atlas segmentation techniques (Sec.~\ref{sec:exp-unified-framework}). We then investigate how the framework can be used - with further modifications of the graph structure - to employ partially annotated atlases for segmentation. This is done using both the slicewise partial annotation strategy (Sec.~\ref{sec:exp-pa-slices}) and scribbles (Sec.~\ref{sec:exp-pa-scribbles}).

The experiments were carried out on two datasets: (1) brain MR images from the ADNI database for hippocampal segmentation (a binary segmentation problem) and (2) cardiac MR images for segmentation of the right and left ventricular cavities and the left ventricle myocardium (i.e. segmentation with multiple labels).

%%%%%%%%%%%%%%%%%%%%%%%%%%%%%%%%%%%%%%%%%%%%%%%%%%%%%%%%%%%%%%%%%%%%%%%%%%%%%%%%%%%%%%%%%%%%%%%%%%%
% *** Evaluation of Proposed Framework for Multi-Atlas Segmentation ***
%%%%%%%%%%%%%%%%%%%%%%%%%%%%%%%%%%%%%%%%%%%%%%%%%%%%%%%%%%%%%%%%%%%%%%%%%%%%%%%%%%%%%%%%%%%%%%%%%%%
\subsection{Evaluation of Proposed Framework for Multi-Atlas Segmentation (MAS)}
\label{sec:exp-unified-framework}

To explore the proposed unifying framework, a number of different configurations were compared which correspond to existing segmentation techniques. To acquire a labelling on a target image, selected atlas images were aligned with the target image using non-rigid registration~\cite{Rueckert1999} and a graph was constructed using each of the chosen configurations. The optimisation proposed in Sec.~\ref{sec:optimisation} was performed to achieve a segmentation result.

The most elementary configuration we studied was multi-atlas segmentation using the majority vote label fusion step (MAS-MV) \cite{Heckemann2006multiatlas,Aljabar2009atlasselection,Rohlfing2004}. For this, we assume a graph structure as shown in Fig.~\ref{fig:multiatlas} and label propagation weights were uniformly set to $\beta_{ij}(x) = 1$. We compared MAS-MV to locally weighted label fusion (MAS-LW) as explored in \cite{Artaechevarria2009,Sabuncu2010,Bai2013}. To this end, we chose propagation weights $\beta_{ij}(x)$ based on a local similarity measure between the target and the atlases as below:
\begin{equation}
\beta_{ij}(x) = K \cdot \exp{  \left( - \frac{ (P_i(x)-P_j(x))^2}{ 2 \pi \sigma_2^2 \cdot |P| } \right) ~,}
\label{eq:lw-measure}
\end{equation}
where $P(x)$ is a patch centred around voxel $x$ and $|P|$ is the patch size. $K$ does not influence the label fusion result and was set to $1$. By modifying the graph configuration to additionally incorporate intra-image edges in the target image, we added a regularisation term as described in Sec.~\ref{sec:regularisation} and shown in Fig.~\ref{fig:multiatlas-reg}. This configuration (further referred to as MASr-LW) implements simultaneous label fusion and regularisation similar to \cite{VanderLijn2008,Lotjonen2010}. It is important to note that these approaches incorporated an additional prior probability term based on intensity models of the data. However, in preliminary experiments, we achieved better results without this term.

\subsubsection{Data and experiment setup} 

The proposed method was applied to 202 images from the ADNI database\,\cite{Jack2008} for which reference segmentations of the hippocampus were made available through ADNI. In a pre-processing step, all images were affinely aligned to the MNI152 template space and intensity-normalised\,\cite{Nyul1999MRScale}. The data were split randomly into two equally sized sets, one for parameter training and one for evaluation. Optimal parameters were chosen for locally weighted label fusion (i.e. the propagaton term) and  for spatial regularisation. The tuning procedure and results are described in Sec.~\ref{sec:param-tuning-exp1}. The terminal connections encoding the data term simply consisted of infinite weights in voxels where manual annotations were available, and zero weight (i.e. missing link) in unlabelled voxels.

\subsubsection{Results}
\label{sec:exp1-results}

\begin{figure}[t]
 \centering
 {\includegraphics[height=.18\textheight]{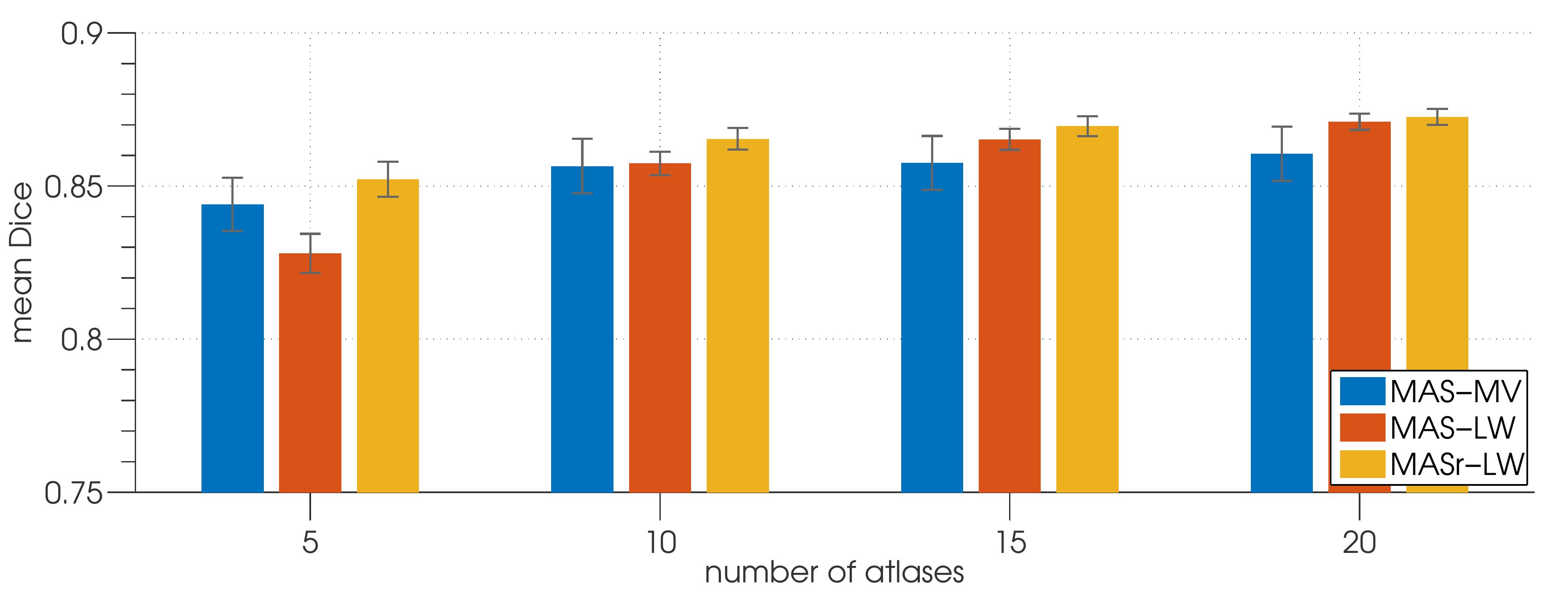}} %\hfill
 \caption{This figure shows mean Dice coefficients for MAS-MV, MAS-LW and MASr-LW using $R=\{5, 10, 15, 20\}$ atlases. The error bars depict the standard error.}
  \label{fig:MAS-eval}
\end{figure}

For evaluation, a $10$-fold cross-validation was performed within the evaluation set. For the each
fold, every test subject was segmented using the training data (i.e. the remaining folds), which
served as the atlas population. This means that for each test subject, the $R$ most similar images
from the remaining folds were used as atlases. Similarity was assessed with normalised mutual information. This was repeated for $R=\{5, 10, 15, 20\}$ to measure the influence of the number of atlases on segmentation accuracy.
Figure~\ref{fig:MAS-eval} shows the mean Dice coefficients of the pooled results. Segmentation results generally increased with the number of atlases used. Majority vote (MAS-MV) was more robust than locally weighted fusion (MAS-LW)  when using $5$ or $10$ atlases, but for larger atlas sets, MAS-LW achieved better results. With additional spatial regularisation, MASr-LW consistently outperformed both MAS-LW and MAS-MV.

%%%%%%%%%%%%%%%%%%%%%%%%%%%%%%%%%%%%%%%%%%%%%%%%%%%%%%%%%%%%%%%%%%%%%%%%%%%%%%%%%%%%%%%%%%%%%%%%%%%
% *** Evaluation of Partial Annotation Strategy A: Slicewise (SW) ***
%%%%%%%%%%%%%%%%%%%%%%%%%%%%%%%%%%%%%%%%%%%%%%%%%%%%%%%%%%%%%%%%%%%%%%%%%%%%%%%%%%%%%%%%%%%%%%%%%%%
\subsection{Evaluation of Partial Annotation Strategy A: Slicewise (PA-SW)}
\label{sec:exp-pa-slices}

This experiment aims to investigate the performance of our framework when using atlas data which were partially annotated through slice-by-slice annotation as proposed in Sec.~\ref{sec:strategy-slices}.

As proposed in Sec.~\ref{sec:segmentation_using_partially_annotated_atlas_data}, we examined two
graph configurations using different propagation schemes. In the first configuration (further
referred to as PA-SW-CONF1) as shown in Fig.~\ref{fig:partial_config_a}, the regularisation term
included spatial regularisation in \emph{all} images (i.e. target and atlases). The propagation term
allowed label propagation from the atlases to the target. In addition, in the second configuration
(further referred to as PA-SW-CONF2), label propagation \emph{between the atlases} was allowed by
expanding the propagation term with inter-atlas connections as shown in
Fig.~\ref{fig:partial_config_b}.

\subsubsection{Data and experiment setup}

The same data was used as in the previous experiment (Sec.~\ref{sec:exp-unified-framework}). To simulate partially annotated atlas data, manual labels of a proportion $q$ of evenly distributed slices in $20$ atlas images were used for segmentation of the target image. 
To determine which slice positions were used, a random offset was determined for each atlas image.
The partial annotations were then transformed to the target space using nonrigid registration~\cite{Rueckert1999}. The data term was built by establishing terminal connections at labelled voxels, while leaving unlabelled voxels unconnected, as explained in Sec.~\ref{sec:data-term-missing-labels}. The proportion of labelled atlas slices ranged from $q=1$ (i.e. fully labelled) to $q=0.1$ (i.e. every 10th slice) to investigate how strongly the atlas label maps could be sub-sampled while achieving robust segmentation results.

The parameters for the propagation term were chosen as in the previous experiment and optimal choices for the regularisation coefficients $a, \sigma_1$ were obtained through parameter tuning as described in detail in Sec.~\ref{sec:param-tuning-exp2}.

\subsubsection{Results}

\begin{figure}[t]
  \centering
  \includegraphics[height=.18\textheight]{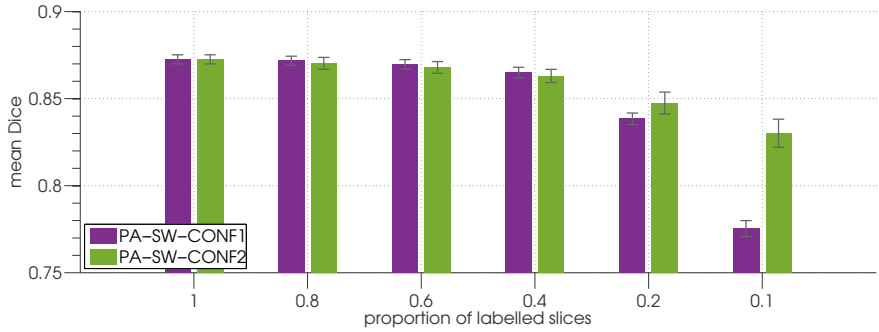} %\hfill
  \caption{This figure shows mean Dice coefficients for slicewise partial annotation (PA-SW) for different proportions $q$ of labelled atlas slices. PA-SW-CONF1 and PA-SW-CONF2 describe the graph configurations and the error bars depict the standard error.}
  \label{fig:PASW-eval}
\end{figure}

\begin{figure}[t]
 \centering
 \subcaptionbox{PA-SW-CONF2 example results at slice position 1\label{fig:pasw-qualresults-good}}
 {\includegraphics[width=\textwidth]{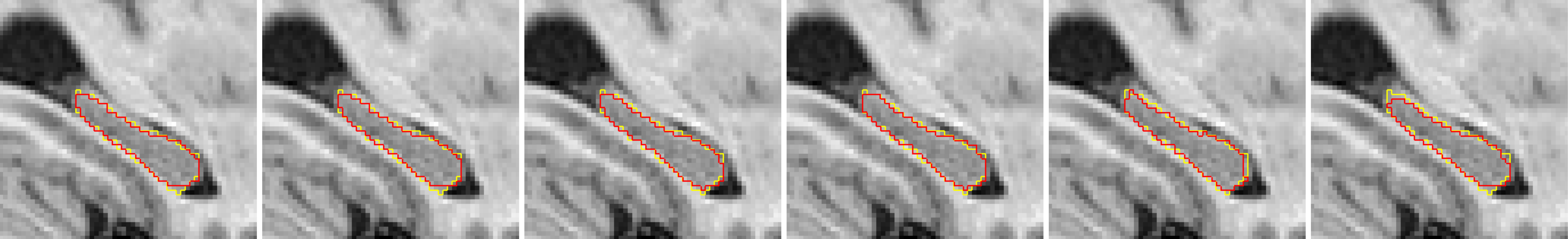}} %\hfill

 \subcaptionbox{PA-SW-CONF2 example results at slice position 2\label{fig:pasw-qualresults-bad}}
 {\includegraphics[width=\textwidth]{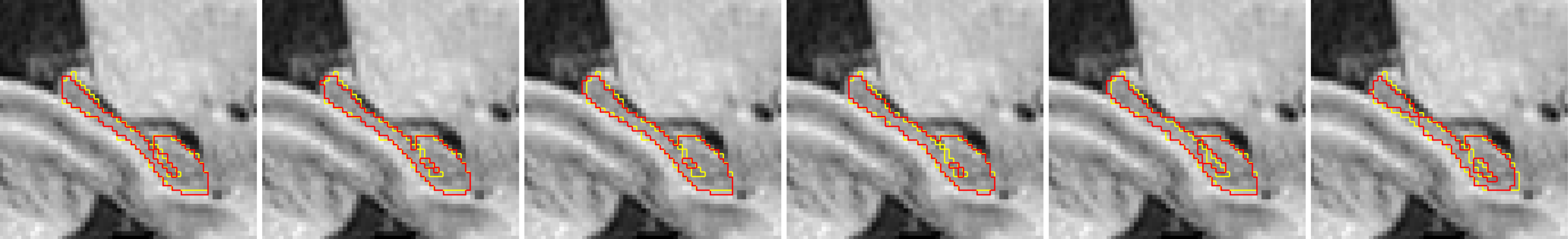}} %\hfill
 \caption{An example segmentation for PA-SW-CONF2 is shown in red, yellow denotes the ground truth segmentation. The same subject is shown at two different slice positions in (a) and (b). From left to right, the proportion of labelled atlas slices $q$ was $1, 0.8, 0.6, 0.4, 0.2, 0.1$. 
 % The chosen subject obtained the median Dice for $q=0.1$.
 }
  \label{fig:pasw-qualresults}
\end{figure}

Results on the evaluation set were obtained using the same 10-fold cross-validation as described in Sec.~\ref{sec:exp1-results}. Figure~\ref{fig:PASW-eval} shows the mean Dice coefficients pooled from all folds for all tested proportions of labelled slices $q$. For $q=1$ (i.e. the group on the left), all atlas slices were labelled. In this case, the proposed graph configurations PA-SW-CONF1 and PA-SW-CONF2 are equivalent to multi-atlas segmentation with regularisation refinement (MASr-LW). It can be seen that reducing the proportion of labelled atlas slices to $q=0.4$ still yields comparable results for both tested configurations. When using fewer labelled slices, the performance decays rapidly for PA-SW-CONF1. For the second configuration CONF2, accuracy decreases as well, but more steadily. However, it is important to remember that the performance trade-off for e.g. $q=0.1$ stems from one tenth of the labelling effort.
Figure~\ref{fig:pasw-qualresults} shows example segmentation results for one subject at two different slice positions (top and bottom rows) for decreasing values of $q$ (left to right). For the slice in Fig.~\ref{fig:pasw-qualresults-good}, even using only every tenth atlas slice (i.e. $q=0.1$ on the very right) did not influence the segmentation result. The slice in Fig.~\ref{fig:pasw-qualresults-bad} was more challenging to segment due to the complex shape of the hippocampus. There, reducing the proportion of labelled atlas slices lead to failure in detecting the folding of the structure. Incorporating constraints preventing holes in the segmentation could potentially help reduce this effect.

%%%%%%%%%%%%%%%%%%%%%%%%%%%%%%%%%%%%%%%%%%%%%%%%%%%%%%%%%%%%%%%%%%%%%%%%%%%%%%%%%%%%%%%%%%%%%%%%%%%
% *** Evaluation of Partial Annotation Strategy B: Scribbles (SC) ***
%%%%%%%%%%%%%%%%%%%%%%%%%%%%%%%%%%%%%%%%%%%%%%%%%%%%%%%%%%%%%%%%%%%%%%%%%%%%%%%%%%%%%%%%%%%%%%%%%%%
\subsection{Evaluation of Partial Annotation Strategy B: Scribbles (PA-SC)}
\label{sec:exp-pa-scribbles}

Finally, we examined the performance of our framework when using data annotated with scribbles as proposed in Sec.~\ref{sec:strategy-scribbles}. 
In a first group of experiments, we investigated the scenario when the scribbles were available only on the atlas images. This partial annotation scenario will be referred to as PA-SC-A and was compared against MASr-LW with fully annotated atlases as a gold standard. 
We used the graph configuration CONF1 (as shown in Fig.~\ref{fig:partial_config_a}) since manual labels were available in roughly the same locations in all images (as opposed to the slicewise annotation strategy where entire slices remained unlabelled). Therefore, the complex propagation scheme CONF2 was not deemed necessary.
In the second group of experiments, we examined scenarios which involve placing scribbles on a target image before automated segmentation, closely related to~\cite{Boykov2000}. In the simplest configuration, scribbled were placed solely on the target image (PA-SC-T)~\cite{Boykov2000}, and no atlases were used. We then investigated if, in addition, a ``scribbled'' atlas database would improve these results (PA-SC-A+T). Here, scribbles were available both in the atlas database and the target image. Lastly, we used fully annotated atlases in combination with a scribbled target image (PA-SC-AF+T) to obtain a target segmentation with the proposed framework.

\subsubsection{Data and experiment setup}

These experiments were performed for multi-label cardiac segmentation. The proposed method was tested on a short-axis cardiac MR (CMR) dataset of 28 subjects in the end-diastole (ED) phase. The CMR data were acquired on a 1.5T Philips Achieva system (Best, The Netherlands) using a 32-channel coil and the balanced-steady state free precession (b-SSFP) sequence. Images in the left ventricular short-axis plane were acquired using the following parameters: $320 \times 320$\,mm field-of-view; 3.0\,ms repetition time (TR); 1.5\,ms echo time (TE); 50\,ms shot duration; 30 cardiac phases; 8\,mm section thickness with a 2\,mm gap. The reconstructed MR images are of dimension $288 \times 288 \times 12$, with voxel spacing $1.23 \times 1.23 \times 10$\,mm. The LV cavity, LV myocardium, and the RV cavity were manually annotated by two experienced imaging scientists. Ten subjects were labelled by one observer, whereas the other 18 were labelled by the second observer. The annotation time for a complete image was approximately 30\,min.

In addition, all images were partially annotated by a third observer. For this purpose, scribbles were placed on every slice for all structures (including the background). The task was set such that the observer should rapidly label large areas while not delineating the structure boundaries. This allowed the annotation time to be reduced to a mean time of $3.9 \pm 0.6$\,min, i.e. a speedup of a factor $> 7$ compared to a full annotation. All manual annotations were done using ITK-SNAP~\cite{Yushkevich2006}.

\begin{figure}[t]
 \centering
 \subcaptionbox{Example image\label{fig:cardiac-data-img}}
 {\includegraphics[height=.15\textheight]{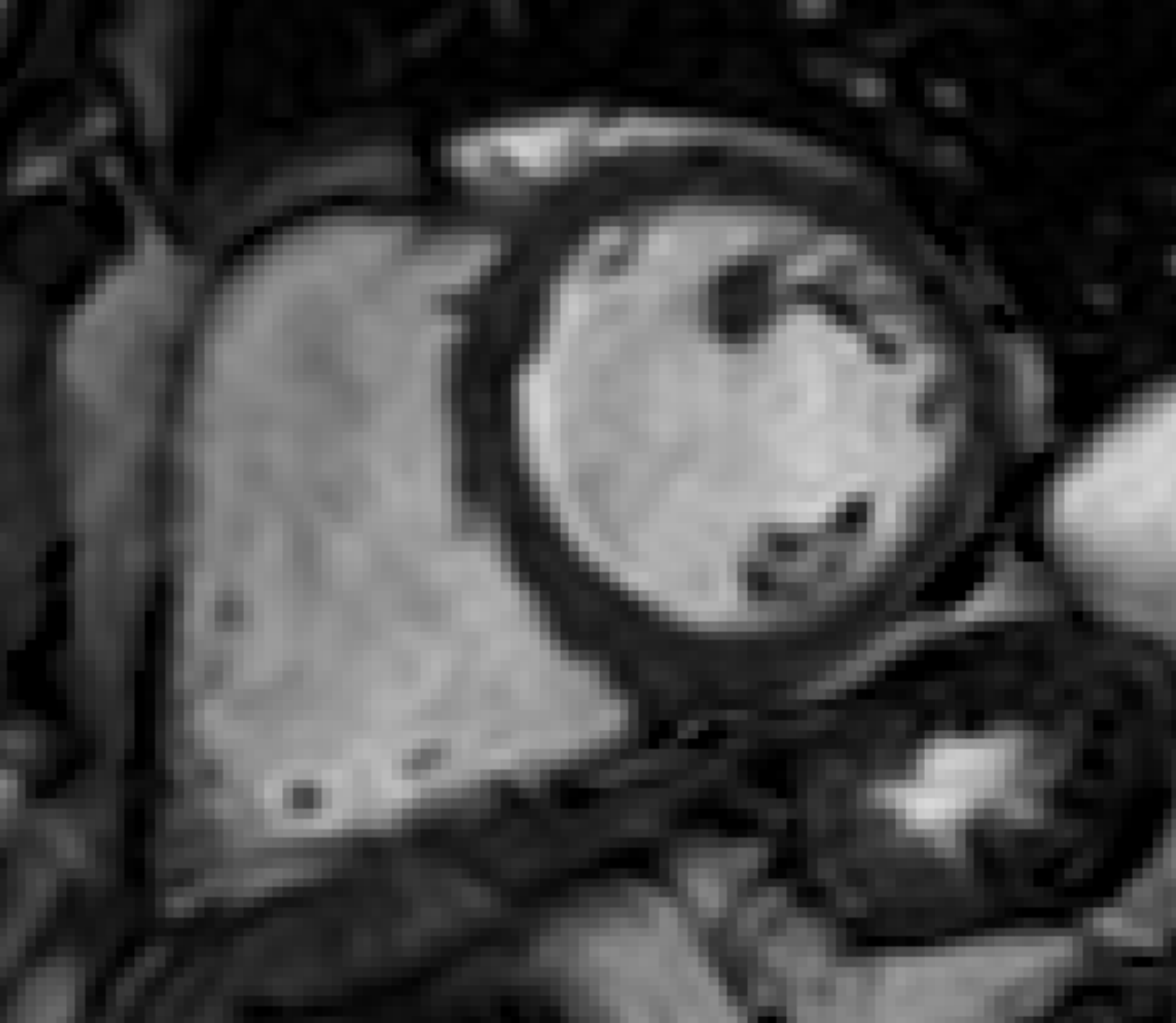}} %\hfill
 \quad%
 \subcaptionbox{Full annotation\label{ffig:cardiac-data-label}}
 {\includegraphics[height=.15\textheight]{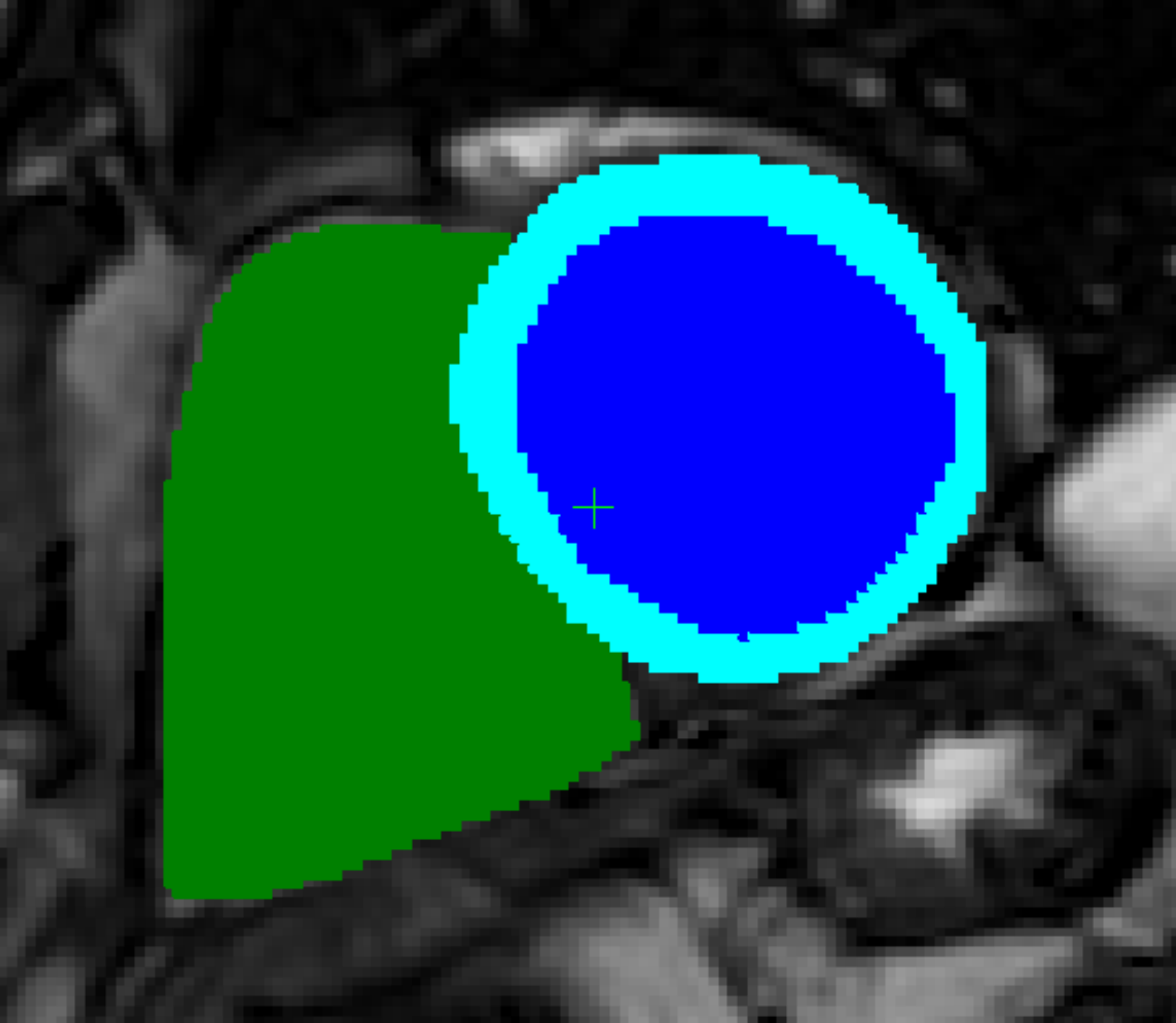}} %\hfill
 \quad%
 \subcaptionbox{Scribbles\label{fig:cardiac-data-scribbles}}
 {\includegraphics[height=.15\textheight]{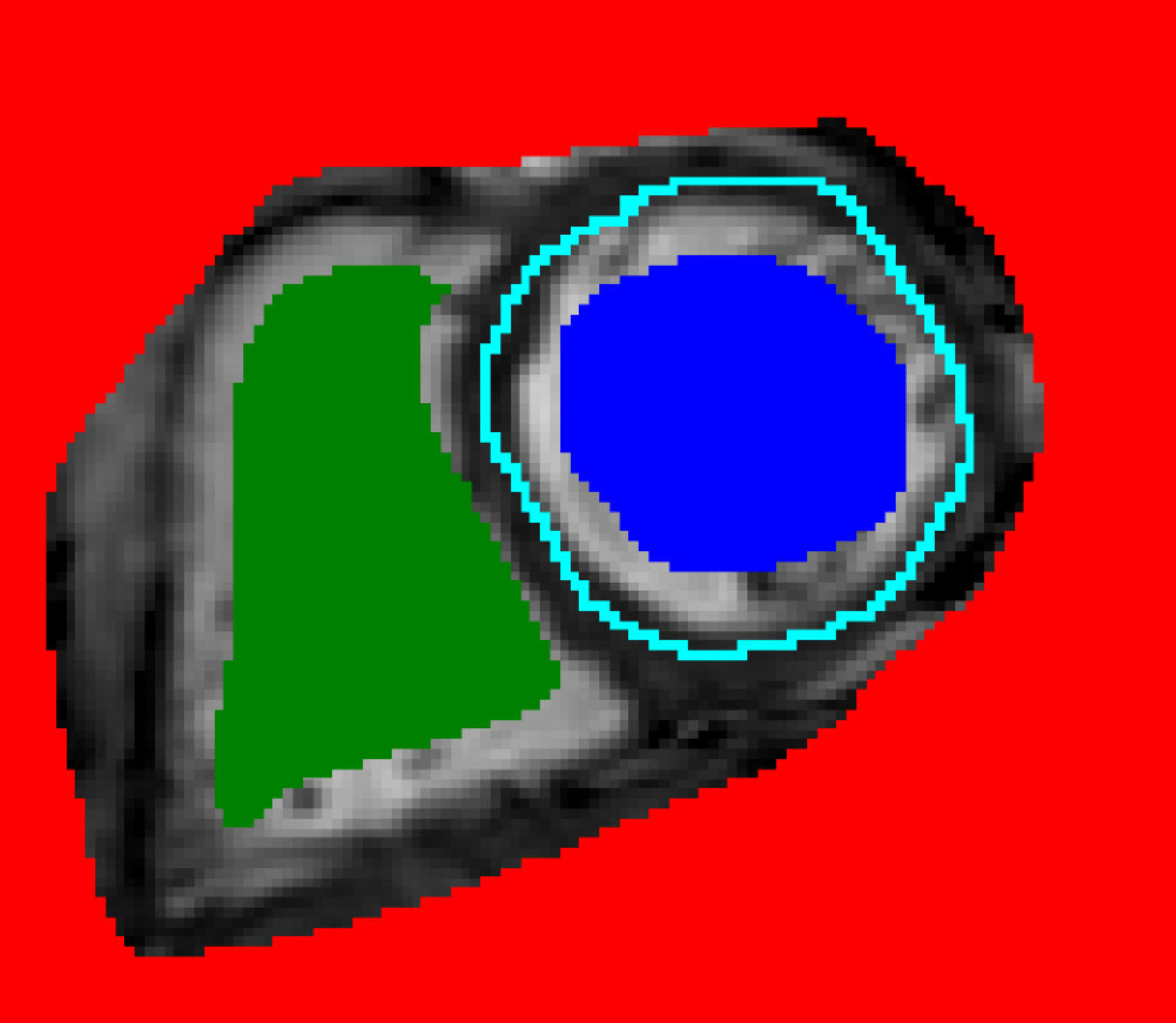}} %\hfill
 \caption{Example cardiac data: (a) shows an image of the heart and (b) shows the complete annotation of the left ventricular cavity (blue), the left ventricular myocardium (cyan) and the right ventricular cavity (green). (c) shows scribbles placed on the same image using ITK-SNAP~\cite{Yushkevich2006}.}
  \label{fig:cardiac-data}
\end{figure}

The propagation weights $\beta_{ij}$ for label fusion were chosen as in \cite{Bai2013}, where the same cardiac dataset was used. There, an exponential kernel was proposed based on the sum of squared distances between two patches centred around corresponding voxels in the target and atlas image. The optimal kernel width was found to be $50$ and the patch size $3 \times 3 \times 1$ voxels. 
Suitable parameters for spatial regularisation $a, \sigma_1$ were found in a training step as described in Sec.~\ref{sec:param-tuning-exp3}.

\subsubsection{Results}

The proposed configurations were evaluated using each image not used during parameter training as a target image. The remaining images were used as atlas images, respectively. For each target subject, the $15$ most similar remaining images were used as atlases as in \cite{Bai2013} (measured with normalised mutual information).

\begin{figure}[t]
 \centering
 \subcaptionbox{Results for experiments using scribbled atlases and MASr-LW\label{fig:res-scribbles-atlases}}
 {\includegraphics[height=.18\textheight]{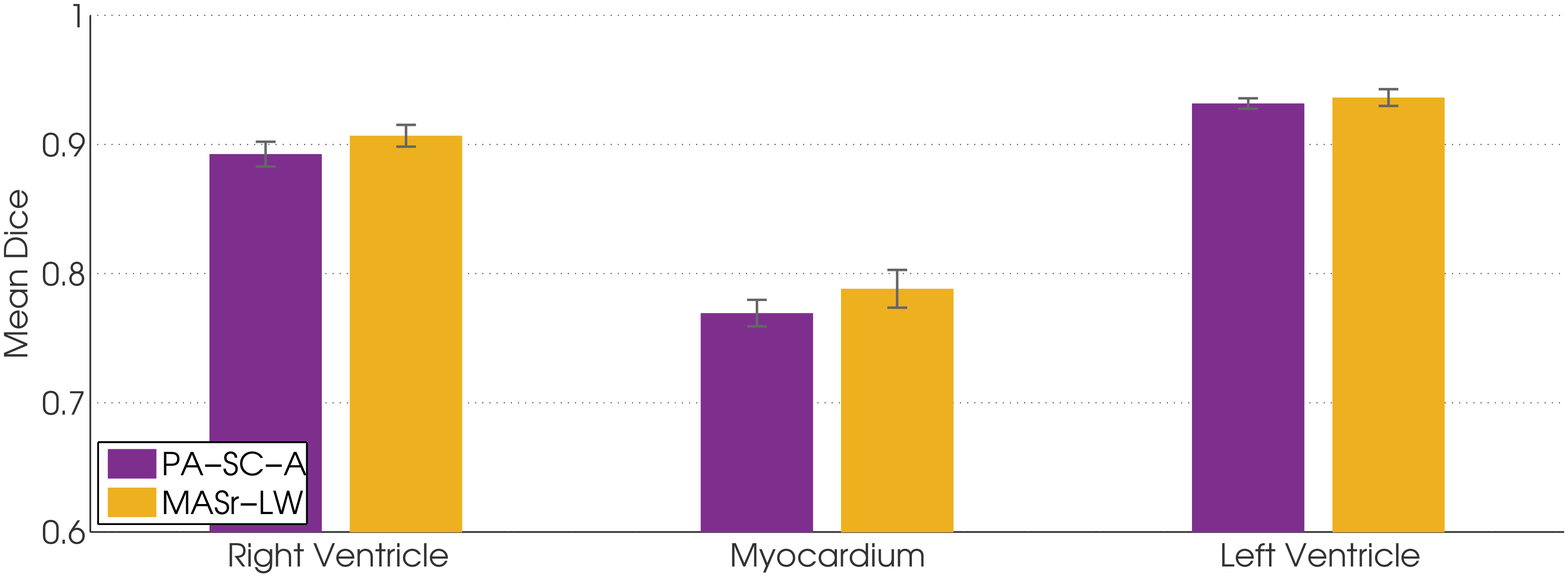}} %\hfill
 % \quad%
 \hfill%
 \subcaptionbox{Results for experiments using scribbled targets\label{fig:res-scribbles-target}}
 {\includegraphics[height=.18\textheight]{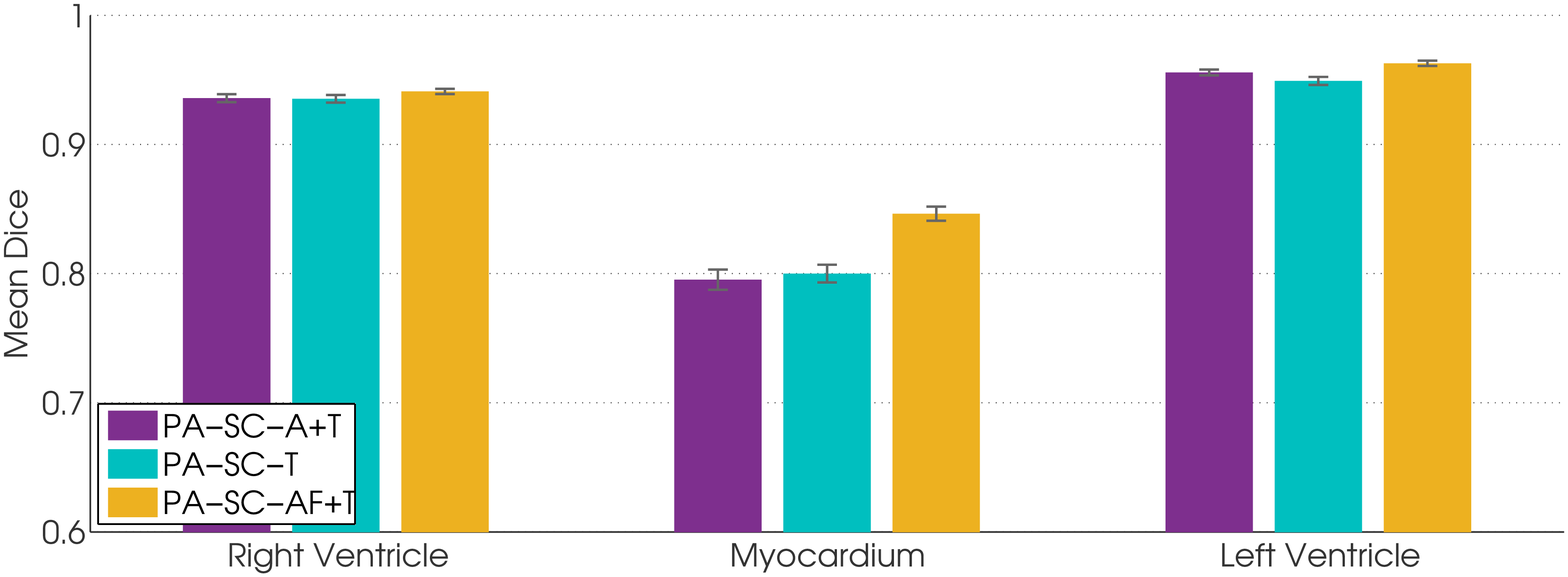}} %\hfill
 \caption{Mean Dice coefficients are shown for all experiments employing scribbles. (a) compares the performance of configurations using scribbled atlas data to fully annotated atlas data and in (b), results are shown for all configurations where the target itself contains scribbles as well.}
  \label{fig:res-scribbles}
\end{figure}

\begin{figure}[t]
  \centering
  \includegraphics[height=.40\textheight]{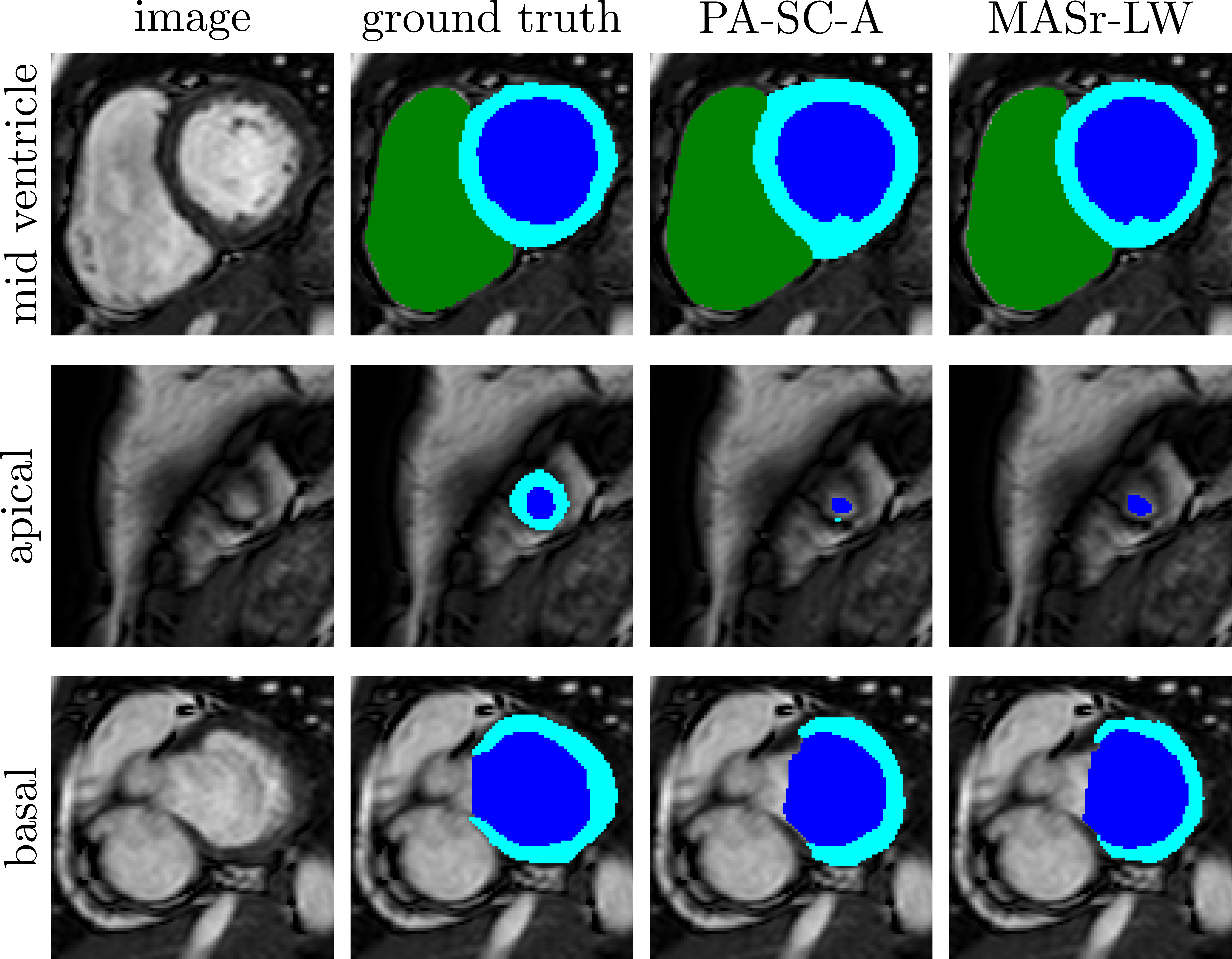} %\hfill
  \caption{Visual results for a mid-ventricular (top), an apical (middle) and a basal slice (bottom) for one subject. The example image, ground truth segmentation, the segmentation obtained with PA-SC-A and MASr-LW are shown from left to right.}
  \label{fig:pasc-qualresults-atlases}
\end{figure}

Figure~\ref{fig:res-scribbles-atlases} shows mean Dice coefficients for the first group of experiments, where scribbles were placed on the atlases, and completely unlabelled target images were segmented using the proposed framework. It can be seen that using scribbled atlases (PA-SC-A) yielded results comparable to MASr-LW (where fully annotated atlases were used) for the right and left ventricle. For the myocardium, using scribbled atlases could not match the accuracy achieved when using fully annotated atlases.
%TODO: statistical significance!
%TODO: how much difference (in Dice)?
Figure~\ref{fig:pasc-qualresults-atlases} shows example segmentation results for one subject. It can be seen that the results of PA-SC-A and MASr-LW are similar. However, since there is no boundary delineation in the scribbled atlases, the resulting segmentation results for PA-SC-A were more intensity driven as can be seen for example in the myocardium in the mid-ventricular view.

\begin{figure}[t]
  \centering
  \includegraphics[height=.40\textheight]{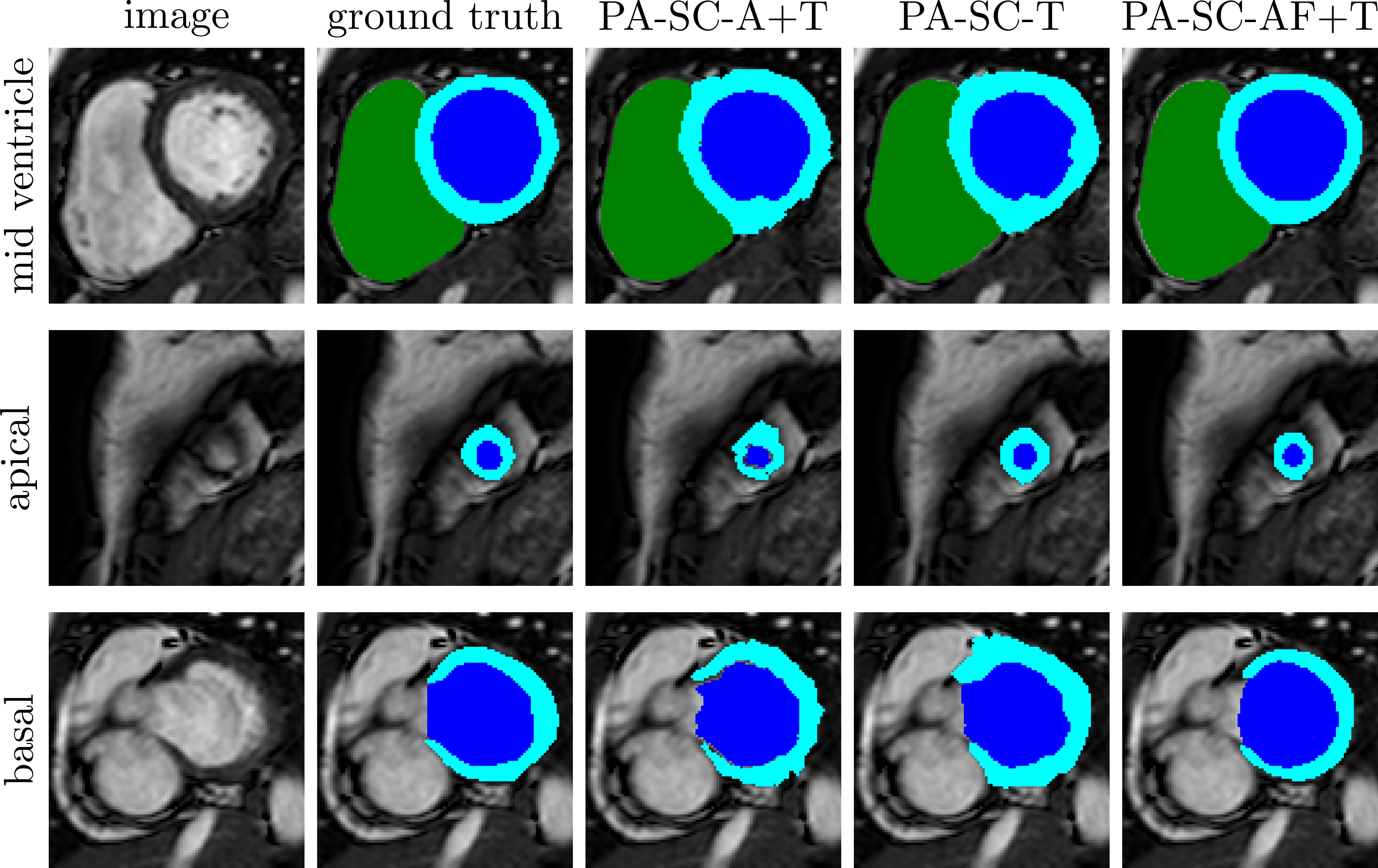} %\hfill
  \caption{Visual results for a mid-ventricular (top), an apical (middle) and a basal slice (bottom) for one subject. The example image, ground truth segmentation, the segmentation obtained with PA-SC-A+T, PA-SC-T, and PA-SC-AF+T are shown from left to right.}
  \label{fig:pasc-qualresults-target}
\end{figure}

The results for the second group of experiments are shown in Fig.~\ref{fig:res-scribbles-target}. Here, the target images to be segmented contained scribbles. In the simplest configuration PA-SC-T, a target segmentation is obtained from the scribbled target image only. Adding the scribbled atlases (PA-SC-A+T) yielded results very similar to PA-SC-T.
%TODO: explain why? explain in numbers?
However, placing scribbles in a target image to aid segmentation using \emph{fully annotated} atlases (PA-SC-AF+T) yielded considerable improvements over both PA-SC-T (as seen in Fig.~\ref{fig:res-scribbles-target}) and MASr-LW (as seen in Fig.~\ref{fig:res-scribbles-atlases}.
%TODO: list numbers!
Visual results for these experiments are shown in Fig.~\ref{fig:pasc-qualresults-target} for the same subject as above. It can be seen that all three methods containing target scribbles were able to detect the myocardium in the apical slice, which was not possible using only atlas information (as seen in the middle row in Fig.~\ref{fig:pasc-qualresults-atlases}). Furthermore, it can be seen that the segmentation obtained with fully annotated atlases and a scribbled target image (PA-SC-AF+T) is visually very similar to the ground truth segmentation, which is also reflected in the high Dice scores reported in Fig.~\ref{fig:res-scribbles-target}.

%%%%%%%%%%%%%%%%%%%%%%%%%%%%%%%%%%%%%%%%%%%%%%%%%%%%%%%%%%%%%%%%%%%%%%%%%%%%%%%%%%%%%%%%%%%%%%%%%%%
% *** Analysis of Parameter Sensitivity ***
%%%%%%%%%%%%%%%%%%%%%%%%%%%%%%%%%%%%%%%%%%%%%%%%%%%%%%%%%%%%%%%%%%%%%%%%%%%%%%%%%%%%%%%%%%%%%%%%%%%
\subsection{Analysis of Parameter Sensitivity}
\label{sec:parameter-tuning}

\subsubsection{Parameter settings for multi-atlas segmentation}
\label{sec:param-tuning-exp1}

\begin{figure}[t]
 \centering
 \includegraphics[height=.17\textheight]{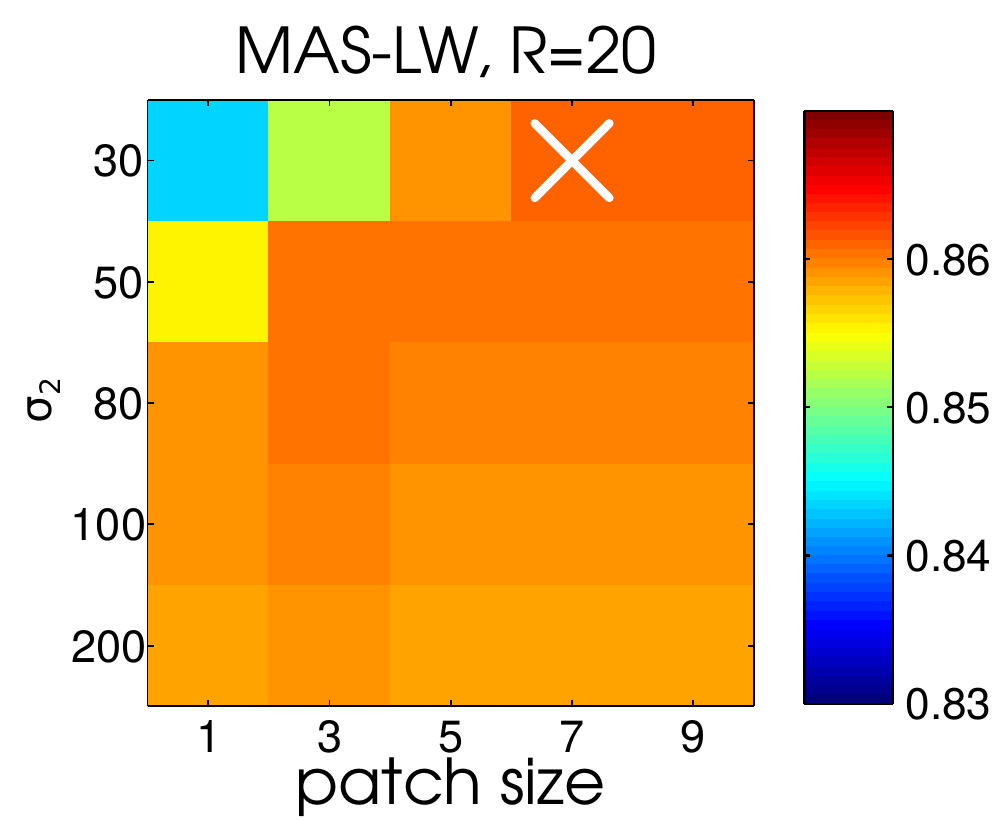} %\hfill
 \caption{This figure shows mean Dice coefficients for a grid search of the parameter choices for MAS-LW on $R=20$ atlases. The white cross marks the optimal parameter choice.}
  \label{fig:MAS-param-training}
\end{figure}

\begin{figure}[t]
 \centering
 \includegraphics[height=.13\textheight]{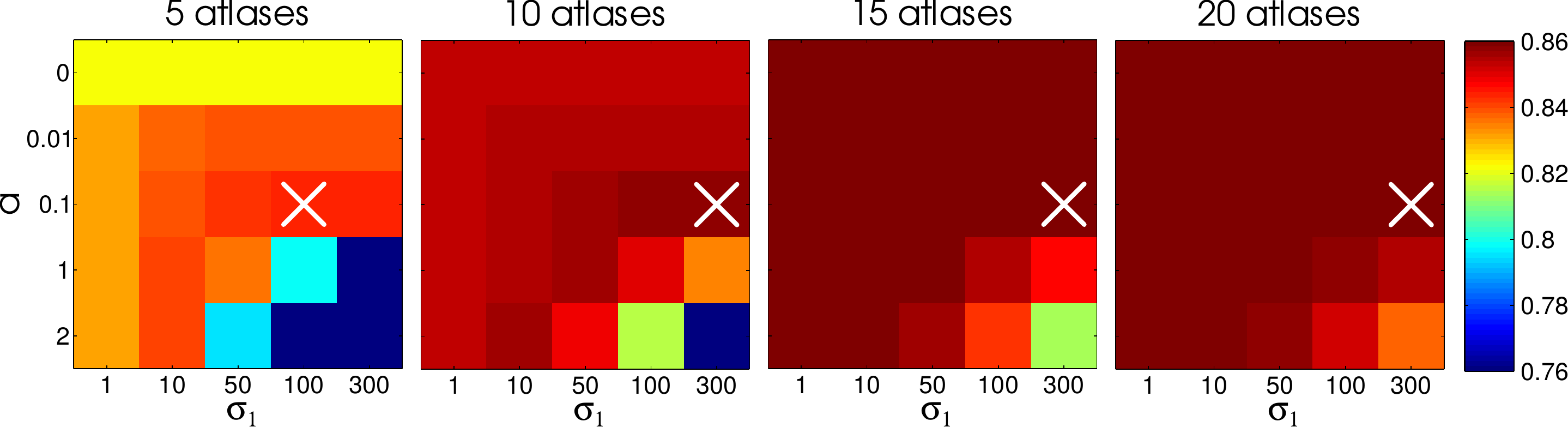} %\hfill
 \caption{This figure shows mean Dice coefficients for a grid search of the parameter choices for MASr-LW using $R=\{5, 10, 15, 20\}$ atlases (left to right). The white cross marks the optimal parameter choice for each experiment.}
  \label{fig:MASr-param-training}
\end{figure}

In this section, we describe the parameter training procedure for the experiments performed in Sec.~\ref{sec:exp-unified-framework}. First, we determined parameter values $\{\sigma_2, |P|\}$  for MAS-LW as introduced in Eq.~\ref{eq:lw-measure}. To do this, $10$ target subjects were randomly drawn from the parameter training data. For each target image, the $20$ most similar images in the remaining training images were used as atlases as recommended in~\cite{Aljabar2009atlasselection} and the segmentation experiments were performed for a parameter range of $|P|=\{ 1, 3, 5, 7, 9 \}$ and $\sigma_2=\{ 30, 50, 80, 100, 200\}$. The parameter set yielding the highest mean Dice coefficient were used for evaluation and subsequent training of the regularisation coefficients $a, \sigma_1$ for MASr-LW. These parameters were trained for $R=\{5, 10, 15, 20\}$ atlases, as we expected the number of atlases to have an influence on the optimal regularisation coefficients. The explored parameter range was $a=\{ 0, 0.01, 0.1, 2\}$ and $\sigma_1=\{ 1, 10, 50, 100, 300\}$.
Figures~\ref{fig:MAS-param-training} and \ref{fig:MASr-param-training} show the results of parameter training.

\subsubsection{Parameter settings for the slicewise (SW) partial annotation strategy}
\label{sec:param-tuning-exp2}

\begin{figure}[ht]
 \centering
 \subcaptionbox{Parameter training for PA-SW-CONF1\label{fig:param-training-PA-SW-CONF1}}
 {\includegraphics[height=.13\textheight]{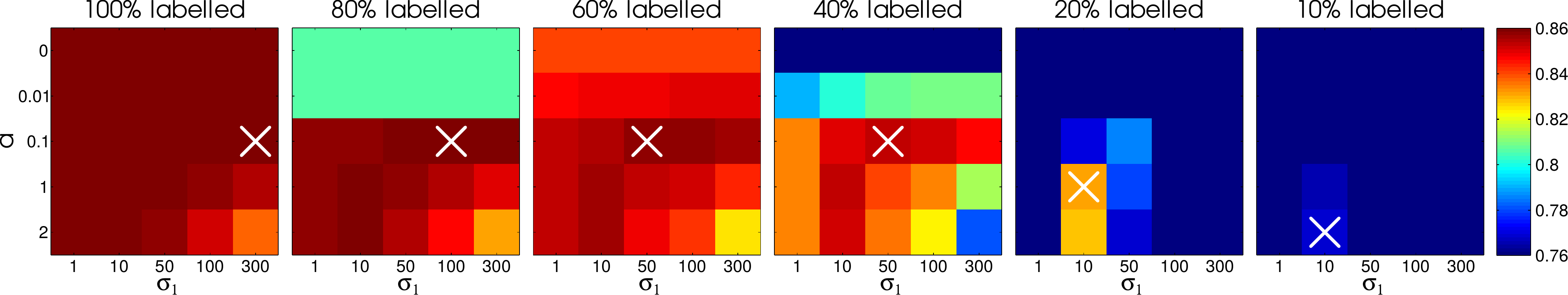}} %\hfill
 \quad\quad\quad%
 \hfill%
 \subcaptionbox{Parameter training for PA-SW-CONF2\label{fig:param-training-PA-SW-CONF2}}
 {\includegraphics[height=.13\textheight]{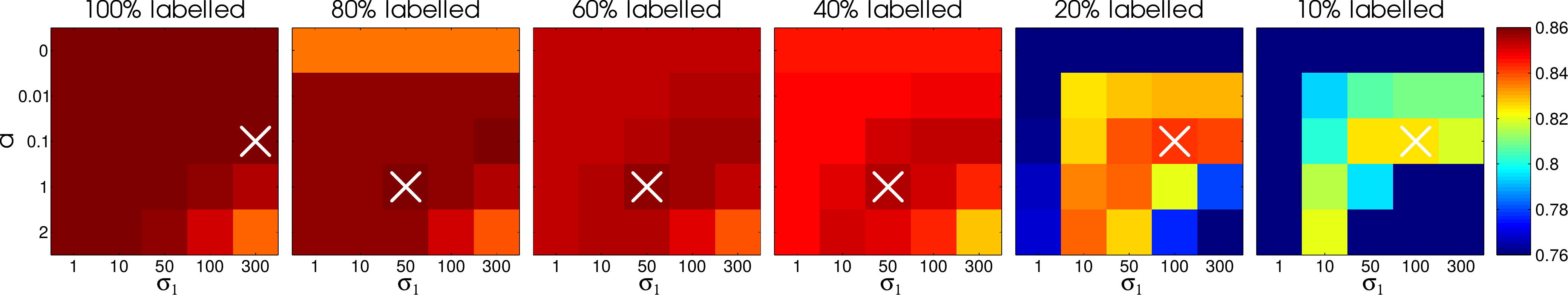}} %\hfill
 \caption{This figure shows mean Dice coefficients for a grid search of the parameter choices using a proportion of $q=\{1, 0.8, 0.6, 0.4, 0.2, 0.1\}$ labelled slices in the atlases (left to right). The white cross marks the optimal parameter choice for each $q$. The colours encode the Dice coefficient (see colorbar on the right). The top (a) and bottom (b) rows show results for CONF1 and CONF2, respectively. }
  \label{fig:PA-SW-param-training}
\end{figure}

For the experiments using slicewise partial annotations (Sec.~\ref{sec:exp-pa-slices}), the spatial regularisation parameters $a, \sigma_1$ were trained on the same training dataset as above. The parameters were tuned separately for both graph examined configurations CONF1 and CONF2. Figure~\ref{fig:PA-SW-param-training} shows optimal parameter
choices for both PA-SW-CONF1 (Fig.~\ref{fig:param-training-PA-SW-CONF1}) and PA-SW-CONF2
(Fig.~\ref{fig:param-training-PA-SW-CONF2}) when using different proportions $q$ of annotated atlas
slices. The parameters with the highest mean Dice score for each configuration and each $q$ were
used during the evaluation.

\subsubsection{Parameter settings for the scribbles (SC) partial annotation strategy}
\label{sec:param-tuning-exp3}

\begin{figure}[t]
 \centering
 \includegraphics[height=.13\textheight]{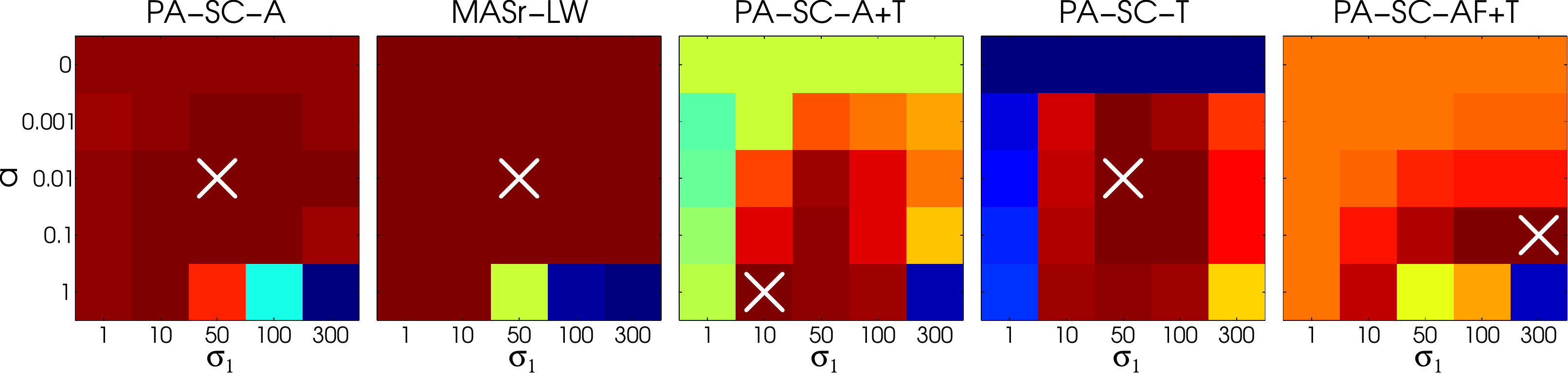} %
 \caption{This figure shows the results of parameter training for all experiments investigating the use of scribbles. The color encodes a measure of combined segmentation accuracy in all structures of interest.}
  \label{fig:PA-SC-param-training}
\end{figure}

Here, parameter training is discussed for the final experiment (Sec.~\ref{sec:exp-pa-scribbles}) where scribbles are used for cardiac segmentation. To find parameter settings for spatial regularisation, $10$ random subjects were selected as target images. For each target subject, the $15$ most similar images from the remaining population were used as atlases as in \cite{Bai2013}. The parameter space was explored on the selected target subjects and the best performing set was used for the remaining population. 
% Thus, a small bias (because of to the incomplete separation from the evaluation set) was allowed due to the limited number of available subjects.
The spatial regularisation parameters $a, \sigma_1$ were explored in a range of $\{0, 0.001, 0.01, 0.1, 1\}$ and $\{1, 10, 50, 100, 300\}$, respectively. Figure~\ref{fig:PA-SC-param-training} shows the training results for all experiment configurations, with optimal parameter choices marked with a white cross.

%%%%%%%%%%%%%%%%%%%%%%%%%%%%%%%%%%%%%%%%%%%%%%%%%%%%%%%%%%%%%%%%%%%%%%%%%%%%%%%%%%%%%%%%%%%%%%%%%%%
%%%%%%%%%%%%%%%%%%%%%%%%%%%%%%%%%%%%%%%%%%%%%%%%%%%%%%%%%%%%%%%%%%%%%%%%%%%%%%%%%%%%%%%%%%%%%%%%%%%
% *** DISCUSSION ***
%%%%%%%%%%%%%%%%%%%%%%%%%%%%%%%%%%%%%%%%%%%%%%%%%%%%%%%%%%%%%%%%%%%%%%%%%%%%%%%%%%%%%%%%%%%%%%%%%%%
%%%%%%%%%%%%%%%%%%%%%%%%%%%%%%%%%%%%%%%%%%%%%%%%%%%%%%%%%%%%%%%%%%%%%%%%%%%%%%%%%%%%%%%%%%%%%%%%%%%

\section{Discussion}
\label{sec:discussion}

In the experiments section, we first demonstrated how our framework can be used to express state-of-the-art techniques through modifications in the graphical representation of the labelling problem (Sec.~\ref{sec:exp-unified-framework}). In particular, label fusion using the majority vote rule~\cite{Heckemann2006multiatlas,Aljabar2009atlasselection} and locally weighted vote rule~\cite{Artaechevarria2009,Sabuncu2010,Bai2013} were compared against locally weighted label fusion with added regularisation for spatial coherence. As expected, using more atlases generally improved segmentation accuracy~\cite{Heckemann2006multiatlas}. The parameters for locally weighted label fusion were only trained using $20$ atlases, which may explain the drop in performance of MAS-LW compared to MAS-MV when using fewer (i.e. $5$ or $10$) atlases. More elaborate parameter training should remove this effect as locally weighted fusion has been shown to outperform majority vote in similar settings~\cite{Sabuncu2010}. Regularisation in the target image (MASr-LW) performed consistently better than MAS-LW. However, improvements became smaller for larger datasets where label fusion from many atlases caused inherent smoothness, yielding decreased benefit from additional spatial regularisation.

By re-interpreting label fusion (i.e. label propagation) as a pairwise component on a Markov Random Field energy function, it is possible to go beyond the scope of existing applications for multi-atlas segmentation. An important point is that the modular graph structure, where pairwise terms can be used for label propagation (between images) or spatial regularisation (within images) and where a unary term can be used to encode manual annotations, allows a relaxation of the annotation requirements for atlases. Therefore, the proposed framework can employ partially annotated images and represent unlabelled voxels simply by removing terminal links in the graph structure. Furthermore, the label propagation and regularisation schemes can be configured in different ways to facilitate information propagation in the graph. In Sec.~\ref{sec:exp-pa-slices}, two configurations were used for hippocampal segmentation using partially labelled atlases where only a proportion of slices in each image were annotated. The results showed that with both configurations, it was possible to achieve robust results when using as little as 40\% of the annotations. Using the configuration where labels were propagated between atlases as well as to the target image (PA-SW-CONF2), it was possible to reduce the amount of labelled slices even further while still obtaining mean Dice coefficients of $0.83 \pm 0.08$ for $q=0.1$. In that case for example, only every tenth slice was labelled in the atlases. Depending on the application, this performance trade-off could be acceptable, and this would mean that partially annotated atlas databases could be built in 10\% of the time required to create a fully labelled dataset. When allowing propagation only between each atlas and the target image (PA-SW-CONF1), the performance decayed as the proportion of labelled atlas slices was reduced. This can be explained by the increased distance between labelled slices, making it more difficult for intra-image regularisation to interpolate labels. In contrast to CONF2, in CONF1 each voxel in the atlases is connected only to its spatial neighbours and the target image. Therefore, there may be large distances (on the graph) between unlabelled and labelled nodes. CONF2 addresses this problem by facilitating propagation between atlases as well, therefore reducing the distances of unlabelled nodes to nodes with strong data terms.

In the slicewise annotation strategy discussed above, the selected slices were completely annotated with detailed delineations of structures of interest. In contrast, scribbles were proposed as an alternative partial annotation strategy in Sec.~\ref{sec:strategy-scribbles}, with the aim to save time by not requiring the observer to delineate the structure boundaries. We chose to design the task such that the scribbled areas were as large as possible without sacrificing speed on annotating details (as shown in Fig.~\ref{fig:cardiac-data-scribbles}). Placing smaller scribbles could further increase speed more, but likely at the expense of segmentation accuracy. The results presented in Fig.~\ref{fig:res-scribbles-atlases} show that using scribbled atlases yielded comparable performance to MASr-LW, albeit with slightly worse accuracy in the myocardium.
The final set of experiments assumed the infrastructure for placing manual scribbles is available at segmentation time, as for example in interactive segmentation~\cite{Boykov2000}. Results (Fig.~\ref{fig:res-scribbles-target}) showed that in this case, the additional help of scribbled atlases did not greatly influence segmentation results, indicating that scribbles in the target directly are sufficient for obtaining an accurate segmentation with the proposed framework. However, it can be seen that in combination with a scribbled target image, a fully annotated atlas set can improve segmentation results considerably in the myocardium, which is the most challenging structure to segment accurately.

%%%%%%%%%%%%%%%%%%%%%%%%%%%%%%%%%%%%%%%%%%%%%%%%%%%%%%%%%%%%%%%%%%%%%%%%%%%%%%%%%%%%%%%%%%%%%%%%%%%
% *** Future Work ***
%%%%%%%%%%%%%%%%%%%%%%%%%%%%%%%%%%%%%%%%%%%%%%%%%%%%%%%%%%%%%%%%%%%%%%%%%%%%%%%%%%%%%%%%%%%%%%%%%%%
\subsection*{Future Work}

In the scope of this paper, the data term was used exclusively to encode manual annotations. However, as briefly described in Sec.~\ref{sec:data-term-missing-labels}, more complex models could be applied to the data term such as intensity models for the structures of interest. Furthermore, it would be of great interest to extend the data term to incorporate weak annotations such as bounding boxes or image tags. Another extension to the proposed framework could move from a voxel-wise representation of the images to a supervoxel representation. This change in the graphical representation could enhance the scalability of the proposed method to larger databases. 

%%%%%%%%%%%%%%%%%%%%%%%%%%%%%%%%%%%%%%%%%%%%%%%%%%%%%%%%%%%%%%%%%%%%%%%%%%%%%%%%%%%%%%%%%%%%%%%%%%%
%%%%%%%%%%%%%%%%%%%%%%%%%%%%%%%%%%%%%%%%%%%%%%%%%%%%%%%%%%%%%%%%%%%%%%%%%%%%%%%%%%%%%%%%%%%%%%%%%%%
% *** CONCLUSION ***
%%%%%%%%%%%%%%%%%%%%%%%%%%%%%%%%%%%%%%%%%%%%%%%%%%%%%%%%%%%%%%%%%%%%%%%%%%%%%%%%%%%%%%%%%%%%%%%%%%%
%%%%%%%%%%%%%%%%%%%%%%%%%%%%%%%%%%%%%%%%%%%%%%%%%%%%%%%%%%%%%%%%%%%%%%%%%%%%%%%%%%%%%%%%%%%%%%%%%%%

\section{Conclusion}
\label{sec:conclusion}

In this paper, we proposed a unifying formulation for label propagation and regularisation based on a novel graphical representation of the labelling problem which is flexible and easily extendable. Small modifications in its configuration allow the use of partially annotated atlas data for segmentation. Experiments on two datasets demonstrated the usefulness of the proposed framework for segmentation using different partial annotation strategies. Pursuing these annotation strategies can save time and make annotating large databases feasible, while leading to robust segmentation results when combined with existing concepts in multi-atlas segmentation.

% % needed in second column of first page if using \IEEEpubid
% %\IEEEpubidadjcol


% \appendices
% \section{Inter-Image Flow Calculation}

% % you can choose not to have a title for an appendix
% % if you want by leaving the argument blank
% \section{}

\ifCLASSOPTIONcompsoc
  \section*{Acknowledgments}
\else
  \section*{Acknowledgment}
\fi

The authors would like to thank Dr. Declan P. O'Regan from MRC Clinical Sciences Centre, Hammersmith Hospital, Imperial College London, for providing the cardiac MR data and the Alzheimer's Disease Neuroimaging Initiative for providing the brain MR data used in this manuscript. The research leading to these results has received funding from the European Union Seventh Framework Programme (FP7/2007 2013) under grant agreement no. 601055, VPH-DARE@IT.

\ifCLASSOPTIONcaptionsoff
  \newpage
\fi

\bibliographystyle{IEEEtran}
\bibliography{IEEEabrv,references_stripped}

\begin{thebibliography}{10}
\providecommand{\url}[1]{#1}
\csname url@samestyle\endcsname
\providecommand{\newblock}{\relax}
\providecommand{\bibinfo}[2]{#2}
\providecommand{\BIBentrySTDinterwordspacing}{\spaceskip=0pt\relax}
\providecommand{\BIBentryALTinterwordstretchfactor}{4}
\providecommand{\BIBentryALTinterwordspacing}{\spaceskip=\fontdimen2\font plus
\BIBentryALTinterwordstretchfactor\fontdimen3\font minus
  \fontdimen4\font\relax}
\providecommand{\BIBforeignlanguage}[2]{{%
\expandafter\ifx\csname l@#1\endcsname\relax
\typeout{** WARNING: IEEEtran.bst: No hyphenation pattern has been}%
\typeout{** loaded for the language `#1'. Using the pattern for}%
\typeout{** the default language instead.}%
\else
\language=\csname l@#1\endcsname
\fi
#2}}
\providecommand{\BIBdecl}{\relax}
\BIBdecl

\bibitem{Jack2008}
C.~R. Jack, M.~Bernstein, N.~C. Fox, P.~Thompson, G.~Alexander, D.~Harvey,
  B.~Borowski, P.~Britson, J.~Whitwell, C.~Ward, A.~Dale, J.~Felmlee,
  J.~Gunter, D.~Hill, R.~Killiany, N.~Schuff, S.~Fox-Bosetti, C.~Lin,
  C.~Studholme, C.~DeCarli, G.~Krueger, H.~Ward, G.~Metzger, K.~Scott,
  R.~Mallozzi, D.~Blezek, J.~Levy, J.~Debbins, A.~Fleisher, M.~Albert,
  R.~Green, G.~Bartzokis, G.~Glover, J.~Mugler, and M.~Weiner, ``{The
  Alzheimer's Disease Neuroimaging Initiative (ADNI): MRI methods},''
  \emph{Magn Reson Im}, vol.~27, no.~4, pp. 685--91, 2008.

\bibitem{Heckemann2006multiatlas}
R.~A. Heckemann, J.~V. Hajnal, P.~Aljabar, D.~Rueckert, and A.~Hammers,
  ``{Automatic anatomical brain MRI segmentation combining label propagation
  and decision fusion},'' \emph{NeuroImage}, vol.~33, no.~1, pp. 115--26, 2006.

\bibitem{Rohlfing2004}
T.~Rohlfing, R.~Brandt, R.~Menzel, and C.~R. Maurer, ``{Evaluation of atlas
  selection strategies for atlas-based image segmentation with application to
  confocal microscopy images of bee brains.}'' \emph{NeuroImage}, vol.~21,
  no.~4, pp. 1428--42, 2004.

\bibitem{Klein2005multiatlas}
A.~Klein, B.~Mensh, S.~Ghosh, J.~Tourville, and J.~Hirsch, ``{Mindboggle:
  automated brain labeling with multiple atlases.}'' \emph{BMC medical
  imaging}, vol.~5, p.~7, 2005.

\bibitem{Iglesias2015a}
J.~E. Iglesias and M.~R. Sabuncu, ``{Multi-Atlas Segmentation of Biomedical
  Images: A Survey},'' \emph{Med Image Anal}, vol.~24, no.~1, pp. 205--219,
  2015.

\bibitem{Wolz2010}
R.~Wolz, P.~Aljabar, J.~V. Hajnal, A.~Hammers, and D.~Rueckert, ``{LEAP:
  learning embeddings for atlas propagation.}'' \emph{NeuroImage}, vol.~49,
  no.~2, pp. 1316--25, 2010.

\bibitem{Artaechevarria2009}
X.~Artaechevarria, A.~Munoz-Barrutia, and C.~Ortiz-de Sol\'{o}rzano,
  ``{Combination strategies in multi-atlas image segmentation: Application to
  brain MR data},'' \emph{IEEE Trans Med Imag}, vol.~28, no.~8, pp. 1266--77,
  2009.

\bibitem{Aljabar2009atlasselection}
P.~Aljabar, R.~A. Heckemann, A.~Hammers, J.~V. Hajnal, and D.~Rueckert,
  ``{Multi-atlas based segmentation of brain images: atlas selection and its
  effect on accuracy.}'' \emph{NeuroImage}, vol.~46, no.~3, pp. 726--38, 2009.

\bibitem{Sabuncu2010}
M.~R. Sabuncu, B.~T.~T. Yeo, K.~{Van Leemput}, B.~Fischl, and P.~Golland, ``{A
  generative model for image segmentation based on label fusion},'' \emph{IEEE
  Trans Med Imag}, vol.~29, no.~10, pp. 1714--29, 2010.

\bibitem{Bai2013}
W.~Bai, W.~Shi, D.~P. O'Regan, T.~Tong, H.~Wang, S.~Jamil-Copley, N.~S. Peters,
  and D.~Rueckert, ``{A probabilistic patch-based label fusion model for
  multi-atlas segmentation with registration refinement: application to cardiac
  MR images.}'' \emph{IEEE Trans Med Imag}, vol.~32, no.~7, pp. 1302--15, 2013.

\bibitem{Iglesias2015}
J.~E. Iglesias, M.~R. Sabuncu, I.~Aganj, P.~Bhatt, C.~Casillas, D.~Salat,
  A.~Boxer, B.~Fischl, and K.~{Van Leemput}, ``{An algorithm for optimal fusion
  of atlases with different labeling protocols},'' \emph{NeuroImage}, vol. 106,
  pp. 451--63, 2015.

\bibitem{Warfield2004}
S.~K. Warfield, K.~H. Zou, and W.~M. Wells, ``{Simultaneous truth and
  performance level estimation (STAPLE): an algorithm for the validation of
  image segmentation.}'' \emph{IEEE Trans Med Imag}, vol.~23, no.~7, pp.
  903--21, 2004.

\bibitem{Wang2012}
H.~Wang, J.~Suh, S.~Das, J.~Pluta, C.~Craige, and P.~Yushkevich, ``{Multi-Atlas
  Segmentation with Joint Label Fusion.}'' \emph{IEEE Trans PAMI}, vol.~35,
  no.~3, pp. 611--23, 2012.

\bibitem{Coupe2011PatchBased}
P.~Coup\'{e}, J.~V. Manj\'{o}n, V.~Fonov, J.~Pruessner, M.~Robles, and D.~L.
  Collins, ``{Patch-based segmentation using expert priors: application to
  hippocampus and ventricle segmentation.}'' \emph{NeuroImage}, vol.~54, no.~2,
  pp. 940--54, 2011.

\bibitem{Rousseau2011}
F.~Rousseau, ``{A supervised patch-based approach for human brain labeling},''
  \emph{IEEE Trans Med Imag}, vol.~30, no.~10, pp. 1852--62, 2011.

\bibitem{VanderLijn2008}
F.~van~der Lijn, T.~den Heijer, M.~Breteler, and W.~J. Niessen, ``{Hippocampus
  segmentation in MR images using atlas registration, voxel classification, and
  graph cuts.}'' \emph{NeuroImage}, vol.~43, no.~4, pp. 708--20, 2008.

\bibitem{Lotjonen2010}
J.~M. L\"{o}tj\"{o}nen, R.~Wolz, J.~R. Koikkalainen, L.~Thurfjell, G.~Waldemar,
  H.~Soininen, and D.~Rueckert, ``{Fast and robust multi-atlas segmentation of
  brain magnetic resonance images.}'' \emph{NeuroImage}, vol.~49, no.~3, pp.
  2352--65, 2010.

\bibitem{Makropoulos2014}
A.~Makropoulos, I.~S. Gousias, C.~Ledig, P.~Aljabar, A.~Serag, J.~V. Hajnal,
  D.~Edwards, S.~J. Counsell, and D.~Rueckert, ``{Automatic whole brain MRI
  segmentation of the developing neonatal brain.}'' \emph{IEEE Trans Med Imag},
  vol.~33, no.~9, pp. 1818--31, 2014.

\bibitem{Ledig2015}
C.~Ledig, R.~a. Heckemann, A.~Hammers, J.~C. Lopez, V.~F.~J. Newcombe,
  A.~Makropoulos, J.~L\"{o}tj\"{o}nen, D.~K. Menon, and D.~Rueckert, ``{Robust
  whole-brain segmentation: application to traumatic brain injury.}'' \emph{Med
  Image Anal}, vol.~21, no.~1, pp. 40--58, 2015.

\bibitem{Rajchl}
M.~Rajchl, J.~S. Baxter, A.~J. McLeod, J.~Yuan, W.~Qiu, T.~M. Peters, and A.~R.
  Khan, ``{Hierarchical max-flow segmentation framework for multi-atlas
  segmentation with Kohonen self-organizing map based Gaussian mixture
  modeling},'' \emph{Med Image Anal}, In press.

\bibitem{Boykov2001}
Y.~Boykov, O.~Veksler, and R.~Zabih, ``{Fast approximate energy minimization
  via graph cuts},'' \emph{IEEE Trans PAMI}, vol.~23, no.~11, pp. 1222--39,
  2001.

\bibitem{Koch2014}
L.~M. Koch, R.~Wright, D.~Vatansever, V.~Kyriakopoulou, C.~Malamateniou, P.~A.
  Patkee, M.~A. Rutherford, J.~V. Hajnal, P.~Aljabar, and D.~Rueckert,
  ``{Graph-Based Label Propagation in Fetal Brain MR Images},'' in
  \emph{MLMI}.\hskip 1em plus 0.5em minus 0.4em\relax Springer, 2014, pp.
  9--16.

\bibitem{kuettel2012imagnet}
D.~Kuettel, M.~Guillaumin, and V.~Ferrari, ``{Segmentation propagation in
  ImageNet},'' in \emph{ECCV}, ser. LNCS, vol. 7578.\hskip 1em plus 0.5em minus
  0.4em\relax Springer, 2012, pp. 459--73.

\bibitem{Rubinstein2012}
M.~Rubinstein, C.~Liu, and W.~T. Freeman, ``{Annotation propagation in large
  image databases via dense image correspondence},'' in \emph{ECCV}, ser. LNCS,
  vol. 7574.\hskip 1em plus 0.5em minus 0.4em\relax Springer, 2012, pp. 85--99.

\bibitem{Cardoso2015}
M.~J. Cardoso, M.~Modat, R.~Wolz, A.~Melbourne, D.~Cash, D.~Rueckert, and
  S.~Ourselin, ``{Geodesic Information Flows: Spatially-Variant Graphs and
  Their Application to Segmentation and Fusion},'' \emph{IEEE Trans Med Imag},
  pp. 1976--88, 2015.

\bibitem{Rother2004}
C.~Rother, V.~Kolmogorov, and A.~Blake, ``{Grabcut: Interactive foreground
  extraction using iterated graph cuts},'' \emph{ACM T Graphic}, vol.~23,
  no.~3, pp. 309--14, 2004.

\bibitem{Chen2014}
L.-C. Chen, S.~Fidler, A.~L. Yuille, and R.~Urtasun, ``{Beat the MTurkers:
  Automatic Image Labeling from Weak 3D Supervision},'' in \emph{CVPR}.\hskip
  1em plus 0.5em minus 0.4em\relax IEEE, 2014, pp. 3198--3205.

\bibitem{Boykov2000}
Y.~Boykov and M.~Jolly, ``{Interactive organ segmentation using graph cuts},''
  in \emph{MICCAI}, 2000, pp. 276--86.

\bibitem{Xu2014}
J.~Xu, A.~G. Schwing, and R.~Urtasun, ``{Tell Me What You See and I will Show
  You Where It Is},'' in \emph{CVPR}.\hskip 1em plus 0.5em minus 0.4em\relax
  IEEE, 2014, pp. 3190--97.

\bibitem{Xu2015}
J.~Xu, A.~G. Schwing, and R.~Urtasun, ``{Learning to Segment Under Various
  Forms of Weak Supervision},'' in \emph{CVPR}.\hskip 1em plus 0.5em minus
  0.4em\relax IEEE, 2015, pp. 3781--90.

\bibitem{Landman2012}
B.~A. Landman, A.~Asman, A.~Scoggins, J.~Bogovic, F.~Xing, and J.~Prince,
  ``{Robust statistical fusion of image labels},'' \emph{IEEE Trans Med Imag},
  vol.~31, no.~2, pp. 512--22, 2012.

\bibitem{Li1994}
S.~Li, ``{Markov random field models in computer vision},'' in \emph{ECCV},
  1994, pp. 361--70.

\bibitem{Yuan2010}
J.~Yuan, E.~Bae, and X.~Tai, ``{A study on continuous max-flow and min-cut
  approaches},'' in \emph{CVPR}, 2010, pp. 2217--24.

\bibitem{Yuan2010a}
J.~Yuan, E.~Bae, X.~Tai, and Y.~Boykov, ``{A continuous max-flow approach to
  potts model},'' in \emph{ECCV}, 2010, pp. 379--92.

\bibitem{Han2011}
D.~Han, J.~Bayouth, Q.~Song, and A.~Taurani, ``{Globally optimal tumor
  segmentation in PET-CT images: A graph-based co-segmentation method},'' in
  \emph{IPMI}, 2011, pp. 245--56.

\bibitem{Trus2014}
W.~Qiu, J.~Yuan, E.~Ukwatta, Y.~Sun, M.~Rajchl, and A.~Fenster, ``{Prostate
  Segmentation: An Efficient Convex Optimization Approach with Axial Symmetry
  Using 3D TRUS and MR Images},'' \emph{IEEE Trans Med Imag}, vol.~33, no.~4,
  pp. 947--60, 2014.

\bibitem{Koch2015}
L.~M. Koch, M.~Rajchl, T.~Tong, J.~Passerat-palmbach, P.~Aljabar, and
  D.~Rueckert, ``{Multi-atlas Segmentation as a Graph Labelling Problem:
  Application to Partially Annotated Atlas Data},'' in \emph{IPMI}, vol. 9123,
  2015, pp. 221--232.

\bibitem{Wolz2013}
R.~Wolz, C.~Chu, and K.~Misawa, ``{Automated abdominal multi-organ segmentation
  with subject-specific atlas generation},'' \emph{IEEE Trans Med Imag},
  vol.~32, no.~9, pp. 1723--1730, 2013.

\bibitem{Wang2014}
Z.~Wang, K.~K. Bhatia, B.~Glocker, A.~Marvao, T.~Dawes, K.~Misawa, K.~Mori, and
  D.~Rueckert, ``{Geodesic patch-based segmentation},'' in \emph{MICCAI}, vol.
  8673, 2014, pp. 666--673.

\bibitem{Kolmogorov2004}
V.~Kolmogorov and R.~Zabin, ``{What energy functions can be minimized via graph
  cuts?}'' \emph{IEEE Trans PAMI}, vol.~26, no.~2, pp. 147--59, 2004.

\bibitem{Chambolle2004}
A.~Chambolle, ``{An algorithm for total variation minimization and
  applications},'' \emph{J Math Imaging Vis}, vol.~20, no.~2, pp. 89--97, 2004.

\bibitem{Rueckert1999}
D.~Rueckert, L.~I. Sonoda, C.~Hayes, D.~L.~G. Hill, M.~O. Leach, and D.~J.
  Hawkes, ``{Nonrigid registration using free-form deformations: application to
  breast MR images},'' \emph{IEEE Trans Med Imag}, vol.~18, no.~8, pp. 712--21,
  1999.

\bibitem{Nyul1999MRScale}
L.~G. Ny\'{u}l and J.~K. Udupa, ``{On standardizing the MR image intensity
  scale.}'' \emph{Magn Reson Med}, vol.~42, no.~6, pp. 1072--81, 1999.

\bibitem{Yushkevich2006}
P.~A. Yushkevich, J.~Piven, H.~C. Hazlett, R.~G. Smith, S.~Ho, J.~C. Gee, and
  G.~Gerig, ``{User-guided 3D active contour segmentation of anatomical
  structures: significantly improved efficiency and reliability},''
  \emph{NeuroImage}, vol.~31, no.~3, pp. 1116--28, 2006.

\end{thebibliography}
%

\end{document}